\documentclass{article}

\usepackage{amsmath,amsfonts,bm}

\def\eqref#1{equation~\ref{#1}}

\def\1{\bm{1}}

\def\rve{{\mathbf{e}}}

\def\rvu{{\mathbf{i}}}

\def\rvu{{\mathbf{u}}}
\def\rvv{{\mathbf{v}}}

\def\rvx{{\mathbf{x}}}

\def\rvz{{\mathbf{z}}}

\DeclareMathAlphabet{\mathsfit}{\encodingdefault}{\sfdefault}{m}{sl}
\SetMathAlphabet{\mathsfit}{bold}{\encodingdefault}{\sfdefault}{bx}{n}

\usepackage{wrapfig}
\usepackage{multirow}
\usepackage{subfiles}
\usepackage{fontawesome}
\usepackage{microtype}
\usepackage{graphicx}
\usepackage{subfigure}
\usepackage{enumitem}
\usepackage{booktabs} 

\usepackage{hyperref}

\usepackage[accepted]{icml2025}

\usepackage{amsmath}
\usepackage{amssymb}
\usepackage{mathtools}
\usepackage{amsthm}
 \usepackage{bbding}
\usepackage[capitalize,noabbrev]{cleveref}

\theoremstyle{plain}
\newtheorem{theorem}{Theorem}[section]

\newtheorem{lemma}[theorem]{Lemma}

\theoremstyle{definition}

\theoremstyle{remark}

\usepackage[textsize=tiny]{todonotes}

\icmltitlerunning{Nonstationary Time Series Forecasting via Unknown Distribution Adaptation}

\begin{document}

\twocolumn[

\icmltitle{Nonstationary Time Series Forecasting via Unknown Distribution Adaptation}



\icmlsetsymbol{equal}{*}

\begin{icmlauthorlist}
	\icmlauthor{Zijian Li}{MBZUAI}
	\icmlauthor{Ruichu Cai}{gdut}
	\icmlauthor{Zhenhui Yang}{gdut}
	\icmlauthor{Haiqin Huang}{gdut}
	\icmlauthor{Guangyi Chen}{cmu,MBZUAI}
	\icmlauthor{Yifan Shen}{MBZUAI}
	\icmlauthor{Zhengming Chen}{gdut}
	\icmlauthor{Xiangchen Song}{cmu}
	\icmlauthor{Zhifeng Hao}{shantou}
	\icmlauthor{Kun Zhang}{cmu,MBZUAI}
\end{icmlauthorlist}

\icmlaffiliation{MBZUAI}{Machine Learning Department, Mohamed bin Zayed University of Artificial Intelligence, United Arab Emirates}
\icmlaffiliation{gdut}{School of Computer Science, Guangdong University, China}
\icmlaffiliation{cmu}{Carnegie Mellon University, USA}
\icmlaffiliation{shantou}{College of Engineering, Shantou University, China }

\icmlcorrespondingauthor{Ruichu Cai}{cairuichu@gmail.com}

\icmlkeywords{Machine Learning, ICML}

\vskip 0.3in
]



\printAffiliationsAndNotice{}  

\begin{abstract}
As environments evolve, temporal distribution shifts can degrade time series forecasting performance. A straightforward solution is to adapt to nonstationary changes while preserving stationary dependencies. Hence some methods disentangle stationary and nonstationary components by assuming uniform distribution shifts, but it is impractical since when the distribution changes is unknown. To address this challenge, we propose the \textbf{U}nknown \textbf{D}istribution \textbf{A}daptation (\textbf{UDA}) model for nonstationary time series forecasting, which detects when distribution shifts occur and disentangles stationary/nonstationary latent variables, thus enabling adaptation to unknown distribution without assuming a uniform distribution shift.
Specifically, under a Hidden Markov assumption
of latent environments, we demonstrate that the latent environments are identifiable. Sequentially, we further disentangle stationary/nonstationary latent variables by leveraging the variability of historical information. 
Based on these theoretical results, we propose a variational autoencoder-based model, which incorporates an autoregressive hidden Markov model to estimate latent environments. Additionally, we further devise the modular prior networks to disentangle stationary/nonstationary latent variables. These two modules realize automatic adaptation and enhance nonstationary forecasting performance. Experimental results on several datasets validate the effectiveness of our approach.

\end{abstract}

\section{Introduction}

Time series forecasting \cite{zhou2021informer,lim2021time,rangapuram2018deep,chatfield2000time,zhang2003time} has achieved pioneering applications in various fields \cite{bi2023accurate,wu2023interpretable,sezer2020financial}. However, the inherent nonstationarity of time series data hinders the forecasting models from generalizing on the temporally varying distribution shift.

Several methodologies are proposed to solve this problem, which can be categorized into two types according to the inter-instance and intra-instance temporal distribution shift assumptions. 
The first type of method assumes that the shift occurs among instances, and each sequence instance is stationary \cite{li2023transferable,oreshkin2021meta}. Therefore, instance normalization \cite{kim2021reversible} or nonstationary attention mechanism \cite{liu2022non} is used to remove nonstationary components and compensate for them in prediction. Another type assumes that the environment changes uniformly \cite{liu2023adaptive,surana2020koopman}. Therefore, some researchers adopt stationarization \cite{virili2000nonstationarity} to remove nonstationarity from time series data. And \cite{liu2022non} partition the time-series data into equally-sized and stationary segments and uses the Fast Fourier Transform to select stationary and nonstationary components. Recent advances \cite{liu2024time} further learn the invariant information to achieve out-of-distribution generalization, but they neglect the environment-related information.
In summary, these methods aim to disentangle the stationary and nonstationary dependency. More discussion about related works can be found in Appendix \ref{app:related_works}.



\begin{figure*}[t]
    \centering
    \label{fig:motivation}
\includegraphics[width=2\columnwidth]{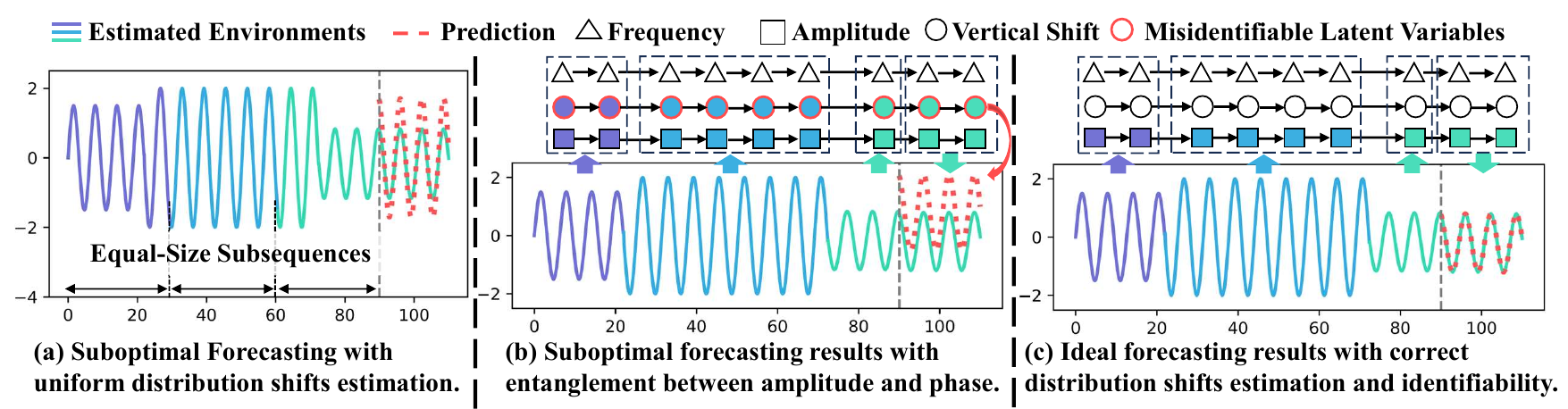}
    \caption{Illustration of nonstationary time series generated from nonstationary amplitude $\square$ (colored graphics) as well as stationary frequency $\triangle$ and vertical shift $\bigcirc$ (white graphics). (a) Methods with a uniform temporal distribution shift assumption cannot disentangle variant and invariant dependencies from the nonstationary segment (green curve), so the prediction with an average amplitude is generated. (b) Even when the latent environments are estimated correctly, the estimated amplitude and vertical shift are entangled, and the vertical shift is considered to change across environments mistakenly, so the upward bias predictions are obtained. (c)With correct environment estimation and latent variable disentanglement, we can achieve ideal forecasting performance. \textit{(Best view in color.)}}
\end{figure*}

Although these methods mitigate the temporal distribution shift to some extent, the assumptions they require are usually too strict since each time series instance or segment may not be stationary, especially when latent environment changes are unknown. Figure \ref{fig:motivation} illustrates an example where a nonstationary sine curve is influenced by the nonstationary latent variable (amplitude) and stationary latent variables (frequency and phase). 
Just like the example in Figure \ref{fig:motivation} (a), existing methods assume a uniform distribution shift and partition the nonstationary time series into three equal segments. However, the purple and green curves remain nonstationary, making disentangling stationary and nonstationary latent variables challenging. 
Besides correct latent environment estimations, proper disentanglement is crucial for adapting to distribution changes. Figure \ref{fig:motivation} (b) shows that if latent variables are entangled—such as mixing amplitude and phase—the model struggles to preserve the stationary dependencies and update the nonstationary ones.

The above points emphasize the importance of a nonstationary time series forecasting model capable of accurately identifying environment changes and disentangling stationary and nonstationary latent variables, as illustrated in Figure \ref{fig:motivation} (c). Standing on this insight, we therefore establish the \textbf{U}nknown \textbf{D}istribution \textbf{A}daptation model (\textbf{UDA} in short), which offers identification guarantees to ensure the performance of nonstationary time series forecasting. Specifically, under the assumption that latent environments follow a Hidden Markov process, we first prove that these environments can be identified from observations. Sequentially, by harnessing the variability of historical information, we can further identify the stationary and nonstationary latent variables, ensuring the proper adaptation to new environments. Guided the theoretical results, we develop the \textbf{UDA} model based on variational autoencoder (VAE). The \textbf{UDA} model is equipped with an autoregressive hidden Markov model for environment estimation and modular prior networks for stationarity/nonstationarity disentanglement. Evaluation of simulation and eight real-world benchmark datasets demonstrates the accuracy of latent environment estimation and identification of latent states, as well as the effectiveness of real-world applications.

\section{Problem Setup}

\subsection{Data Generation Process for Time Series Data}

\begin{figure}
    \centering
\includegraphics[width=0.8\columnwidth]{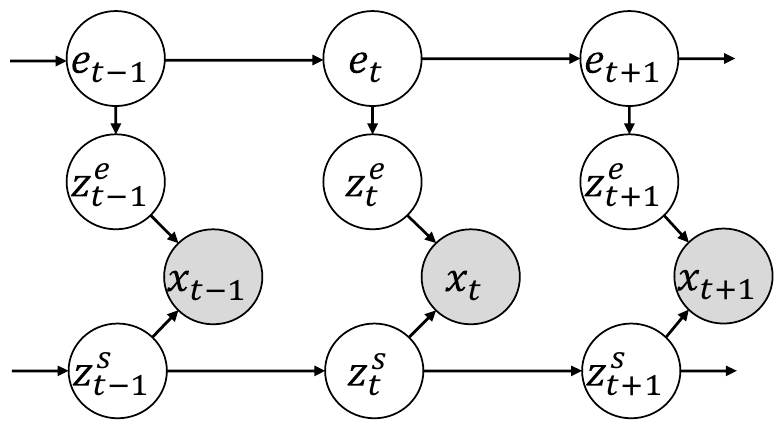} 
\vspace{-0.5cm}
	\caption{Data generation process of nonstationary time series. $\rve_{t}$ denote the discrete latent environment variables, $\rvz_{t}^e$ denote the nonstationary latent variables, and $\rvz_{t}^s$ denote the stationary latent variables. We assume that the number of $\rve$ is known, but when the temporal distribution shift occurs is unknown.} 
    \vspace{-2mm}
    \label{fig:data_generation1}
\end{figure}

To illustrate how we address the nonstationary time series forecasting problem, 
we begin with the data generation process as shown in Figure \ref{fig:data_generation1}. Suppose that we have time series data with discrete time steps, $\textbf{X}=\{\rvx_1, \rvx_2,\cdots, \rvx_T\}$, where $\rvx_t \in \mathbb{R}$ are generated from latent variables $\rvz_t \in \mathcal{Z} \subseteq \mathbb{R}^n$ by an invertible and non-linear mixing function $g$ as shown in Equation (\ref{equ:data_gen1})
\begin{equation}
\label{equ:data_gen1}
\rvx_t = g(\rvz_t).
\end{equation}
Note that $\rvz_t$ are divided into two parts, i.e., $\rvz_t=\{\rvz^s_t, \rvz^e_t\}$, where $\rvz^s_t \in \mathbb{R}^{n_s}$ denote the environment-irrelated stationary latent variables, $\rvz^e_t \in \mathbb{R}^{n_e}$ denote the environment-related nonstationary latent variables, and $n_e+n_s=n$. Specifically, the $i$-th dimension stationary latent variable $z_{t,i}^s$ is time-delayed and causally related to the historical stationary latent variables $\rvz_{t-\tau}^s$ with the time lag of $\tau$ via a nonparametric function $f_i^s$, which is formalized as follows:
\begin{equation}
\label{equ:data_gen2}
z_{t,i}^s\!=\!f^s_i(\{z^s_{t-\tau,k}|z^s_{t-\tau,k}\!\in\! \mathbf{Pa}(z_{t,i}^s)\}, \varepsilon^s_{t,i})\ \ \text{with} \ \ \varepsilon^s_{t,i} \sim p_{\varepsilon_i^s},
\end{equation}
where $\mathbf{Pa}(z_{t, i}^s)$ denotes the set of latent variables that directly cause $\rvz^s_{t, i}$ and $\varepsilon^s_{t, i}$ denotes the temporally and spatially independent noise extracted from a distribution $p_{\varepsilon_i^s}$. Moreover, the nonstationary latent variables $\rvz^e_t$ are influenced by the latent and discrete environment variables $\rve_t$, which follow a first-order Markov process with $\bm{\mathrm{E}} \times \bm{\mathrm{E}}$ transition matrix $\bm{\mathrm{A}}$ and $\bm{\mathrm{E}}$ is the cardinality of $\rve_t$. More specifically, we let the $(k,l)$-th entry $\bm{\mathrm{A}}_{k,l}$ be the probability from the state $k$ to the state $l$. As a result, the generation process of the $j$-th dimension nonstationary latent variable $\rvz_{t,j}^e$ can be formalized as:
\begin{equation}  
\label{equ:data_gen3}
\left\{\begin{array}{l}  \rve_1, \rve_2, \cdots, \rve_T\sim \text{Markov Chain}(\mathbf{A})
\\z_{t,j}^e=f^e_j(\rve_t, \varepsilon^e_{t,j})\ \ \text{with} \ \ \varepsilon^e_{t,j} \sim p_{\varepsilon_j^e} ,
\end{array}\right.
\end{equation} 
in which $f_j^e$ is a bijection function and $\varepsilon^e_{t,j}$ is the mutually-independent noise extracted from $p_{\varepsilon_j^e}$.
To better understand the data generation process in Figure \ref{fig:data_generation1}, we provide a comprehensible example of human driving. First, we let $\rvx_t$ be the speed of a car. Then
$\rve_t$ denotes the action of the driver, i.e., speeding or braking, and $\rvz^e_t$ denotes the engine power or 
acceleration. Finally, $\rvz_t^s$ denotes road conditions such as flat and slippery roads, which are irrelated to the actions.

\subsection{Identifying Distribution of Time Series Data for Nonstationary Time Series Forecasting}
Based on this data generation process, we aim to address the nonstationary time series forecasting problem, i.e., to predict the future observation $\{\rvx_{t+1}, \rvx_{t+2}, \cdots, \rvx_T\}$ only from the historical observation data $\{\rvx_1, \rvx_2, \cdots, \rvx_t\}$. 
Mathematically, our \textit{goal} is to identify the joint distribution of the historical and future time series data. By combining the generation mechanism of Fig. \ref{fig:data_generation1}, the joint distribution can be further derived as follows:
\begin{equation}
\begin{split}
p(\textbf{X})&=\sum_{\rve}\int_{\rvz^e}\int_{\rvz^s}p(\textbf{X}, \rve, \rvz^e, \rvz^s)d_{\rvz^e}d_{\rvz^s}\\&= \sum_{\rve}\int_{\rvz^e}\int_{\rvz^s}p(\textbf{X}|\rvz^e,\rvz^s)p(\rvz^e|\rve)p(\rve)p(\rvz^s) d_{\rvz^e} d_{\rvz^s},
\end{split}
\vspace{-3mm}
\end{equation}
where $\rve:=\{\rve_1,\cdots, \rve_T\}, \rvz^e:=\{\rvz^e_1,\cdots, \rvz^e_T\}$, and $\rvz^s:=\{\rvz^s_1,\cdots, \rvz^s_T\}$ (we omit the subscripts due to limited space). Therefore, the joint distribution is determined by modeling the following four distributions: 1) the generative model of observations given stationary and nonstationary latent variables, i.e., $p(\rvx|\rvz^e,\rvz^s)$; 2) the marginal distribution of latent environment variables, i.e., $p(\rve)$;
3) the distribution of stationary latent variables,i.e., $p(\rvz^s)$; and 4) the conditional distribution of nonstationary latent variables, i.e., $p(\rvz^e|\rve)$.

\section{Identification of Latent Variables}
In this section, we show the identifiability \footnote{Please refere to Appendix \ref{app:identification} for the definition of different types of identification.} of these latent
variables. To well establish the identification results of latent variables, we first leverage Theorem \ref{the:block} to show that the stationary and nonstationary latent variables are block-wise identifiable by employing the fact that $\rvz^e_{t+1}$ depend on $\rvz_{t-1}^e$ given $\rvx_t$. Sequentially, we show that the transition of the latent environments is identifiable up to label swapping, which is shown in Lemma \ref{the:env}. Finally, we prove that stationary and nonstationary latent variables are component-wise identifiable, which is shown in the Lemma \ref{the:component}. 

\subsection{Latent Environment Identification}
To partition the stationary latent variables $\rvz_t^s$ and nonstationary latent variables $\rvz_t^e$, we propose the block-wise identification theory, which is shown in Theorem \ref{the:block}.

\begin{theorem}
\label{the:block}
(\textbf{Block-wise identifiability of the nonstationary latent variables $\rvz_t^e$ and the stationary latent variables $\rvz_t^s$.}) We follow the data generation process in Figure 2 and Equation (1)-(3), then we make the following assumptions:
\begin{itemize}[itemsep=2pt,topsep=0pt,parsep=0pt,leftmargin=0.3cm]
    \item A1 
    (\textbf{Smooth and Positive Density:}) The probability density function of latent variables is smooth and positive, i.e., $p(\rvz^e_t|\rvz_{t-1}^e,\rvz_{t-2}^e)>0$ over $\mathcal{Z}_t^e,\mathcal{Z}_{t-1}^e$ and $\mathcal{Z}_{t-2}^e$.
    \item A2 (\textbf{Sufficient Variability of Historical Information:}) For any $\rvz_t^e\in\mathcal{Z}_t^e\subseteq \mathbb{R}^{n_e}$, $\bar{\rvv}_{t-1,1},\cdots,\bar{\rvv}_{t-1,n_e}$ as $n_e$ vector functions in $z_{t-2,1},\cdots,z_{t-2,l},\cdots,z_{t-2,n_e}$ are linear independent, where $\bar{\rvv}_{t-2,l}$ are formalized as follows:
    \begin{equation}
        \bar{\rvv}_{t-2,l}=\frac{\partial^2 \log p(\rvz_t^e|\rvz_{t-1}^e,\rvz_{t-2}^e)}{\partial z_{t,k}^e\partial z_{t-2,l}^s}
    \end{equation}
    \item A3 \textbf{(Sufficient Variability of Environments:}) There exist two values of $\rvu=\{\rvz_{t-1}^e,\rvz_{t-2}^e\}$, i.e., $\rvu_1$ and $\rvu_2$, s.t., for any set $\mathcal{A}_{\rvz_t}\subseteq \mathcal{Z}_t$ with non-zero probability measure and $\mathcal{A}_{\rvz_t}$ cannot be expressed as $B_{\rvz_t^s}\times \mathcal{Z}_t^e$, for any $B_{\rvz_t^s} \subset \mathcal{Z}_t^s$, we have:
    \begin{equation}
        \int_{\rvz_t\in A_{\rvz_t}}p(\rvz_t|\rvu_1)d \rvz_t\neq \int_{\rvz_t\in A_{\rvz_t}}p(\rvz_t|\rvu_2)d \rvz_t
    \end{equation}
\end{itemize}
Then, by learning the data generation process, $\rvz_t^e$ and $\rvz_t^s$ are block-wise identifiable.
\end{theorem}

\textbf{Proof Sketch.} The proof can be found in Appendix \ref{the:blockwise} First, we construct an invertible transformation $h$ between the ground-truth latent variables $\rvz_t$ and the estimated ones $\hat{\rvz}_t$. According to the data generation process in Figure \ref{fig:data_generation1}, we find that $\rvz_t^e$ is dependent on $\rvz_{t-2}$ while $\rvz_t^s$ is independent of $\rvz_{t-2}$ given $\rvz_{t-1}$. Hence we can construct a full-rank linear system, where the only solution of $\frac{\partial \rvz^e_t}{\partial \hat{\rvz}_t^s}$ is zero. Because of the invertibility of the Jacobian of $h$ and the variability of historical information, both $\rvz_t^e$ and $\rvz_t^s$ are block-wise identifiable.



Based on Theorem \ref{the:block}, we can make sure that the estimated $\hat{\rvz}_t^e$ contain all the information of the truth $\rvz_t^e$ and do not contain the information of stationary latent variables $\rvz_t^s$. So we can consider $\rvz_t^e$ as observed variables and further leverage the results from \cite{allman2009identifiability}, as shown in Lemma \ref{the:env}.


\begin{lemma}
\label{the:env}
\textbf{(Identifiability of the latent environment $\rve_t.$ \cite{allman2009identifiability})} Suppose the observed data are generated following the data
generation process in Figure 3 and Equation (1)-(3). Then we further make the following assumptions:
\begin{itemize}[itemsep=2pt,topsep=0pt,parsep=0pt,leftmargin=0.3cm]
    \item A4 (\underline{\textit{Prior Environment Number:}}) The number of latent environments, $E$, is known.
    \item A5 (\underline{\textit{Full Rank:}}) The transition matrix $\mathbf{A}$ is full rank.
    \item A6 (\underline{\textit{Linear Independence:}}) For $e=1,2,\cdots, E$, the probability measures $\mu_e=p(\rvz_t^e|e_t)$ are linearly independence and for any two different probability measures $\mu_i, \mu_j$, their ratio $\frac{\mu_{i}}{\mu_{j}}$ are linearly independence.
\end{itemize}
Then, by modeling the observations $\rvx_1,\cdots, \rvx_t$, the joint distribution of the corresponding latent environment variables $p(\rve_1,  \cdots, \rve_t)$ is identifiable up to label swapping of the hidden environment.
\end{lemma}
\textbf{Proof Sketch.} First, given any three consecutive observations $\rvx_1, \rvx_2, \rvx_3$ with the corresponding latent environments $\rve_1, \rve_2, \rve_3$, we derive the joint distribution of $p(\rvx_1, \rvx_2, \rvx_3)$ to the product of three independent measures w.r.t. $p(\rve_2)$. Sequentially, by employing the extension of Kruskal's theorem \cite{kruskal1977three,kruskal1976more}, the latent environment variables can be detected with identification guarantees. The detailed proof of Lemma \ref{the:env} is provided in Appendix \ref{app:the1}. 

\subsection{Component-wise Identification of Stationary and Nonstationary Latent Variables}
Based on the aforementioned theoretical results, we prove that the stationary and nonstationary latent variables are component-wise identifiable with the help of nonlinear ICA.

\begin{lemma}
\label{the:component}
(\textbf{Component-wise Identification of the stationary latent variables $\rvz^s_t$ and nonstationary latent variables $\rvz_t^e$.}\cite{yao2021learning}) Following the data generation process in Figure \ref{fig:data_generation1}, smooth and density as well as the similar sufficient variability assumptions of Theorem \ref{the:block}, we further assume that the latent variables are conditionally independent, $\rvz^s_t$ is component-wise identifiable.
\end{lemma}

\textbf{Proof Sketch} The proof can be found in the Appendix \ref{app:the2}, which contains the identification of the stationary and nonstationary variables, respectively. The proof of both types of latent variables is similar. Specifically, we first construct an invertible transformation $h$ between the ground-truth latent variables and estimated ones. Then we employ the variance of different environments to construct a full-rank linear system, where the only solution is zero. 


\subsection{Comparison with Existing Methods}
\begin{table}[]
\vspace{-3mm}
\caption{Attributes of causal representation learning theories. A check denotes that a method has an attribute or can be applied to a setting, whereas a cross denotes the opposite.}
\label{tab:compare}
\resizebox{0.5\textwidth}{!}{%
\begin{tabular}{@{}c|ccc@{}}
\toprule
Methods & \begin{tabular}[c]{@{}c@{}}Partitioned\\  Subspace\end{tabular} & \begin{tabular}[c]{@{}c@{}}Time-Delayed \\ Causal Relations\end{tabular} & \begin{tabular}[c]{@{}c@{}}No Extra Assumptions \\ on Transition\end{tabular} \\ \midrule
HMNLICA  & \XSolidBrush                                                                      & \XSolidBrush                                                                         & \XSolidBrush                                                                             \\
NCTRL   & \XSolidBrush                                                                      & \CheckmarkBold                                                                         & \XSolidBrush                                                                             \\
CtrlNS  & \XSolidBrush                                                                      & \CheckmarkBold                                                                         & \XSolidBrush                                                                             \\
UDA     & \CheckmarkBold                                                                      & \CheckmarkBold                                                                         & \CheckmarkBold                                                                             \\ \bottomrule
\end{tabular}%
}
\vspace{-8mm}
\end{table}

Although recent advances \cite{song2024causal,song2023temporally,halva2020hidden} also achieve the identification for temporal representation under unknown nonstationarity, our method works under less restrictive conditions and better reflects real-world scenarios, as summarized in Table \ref{tab:compare}.

First, compared to existing methods, the proposed UDA is better suited for nonstationary time series forecasting, as it allows for partitioned subspaces. In contrast, methods like \citet{halva2020hidden} cannot be applied to the data generation process, as shown in Figure \ref{fig:data_generation1}, because the causal relationships induced by stationary latent variables disrupt the conditional independence among nonstationary latent variables. Moreover, our method allows for time-delayed causal relationships among latent variables, specifically reflected in stationary latent variables. 

Most importantly, compared with other methods, our approach does not impose additional assumptions on transitions among latent variables. For instance, \citet{halva2020hidden} assume the absence of time-delayed causal relationships, and \citet{song2023temporally} constrains transitions to a nonlinear Gaussian family with unique indexing \cite{balsells2023identifiability}. Furthermore, \citet{song2024causal} further assume a sparse latent transition. These assumptions may not be met in real-world scenarios.

\subsection{Discussion of Assumptions}
For a better understanding of our theoretical results, we further provide detailed explanations and implications of the assumptions of these theories, as well as how they relate to the real-world time series data.

\textbf{Smooth, Positive, and Conditional Independent Density.} This assumption is commonly used in existing identification results \cite{yao2022temporally,yao2021learning}. In real-world scenarios, a smooth and positive density implies continuous changes in historical information, such as temperature fluctuations in weather data. To achieve this, collecting a large amount of data is essential for accurately learning transition probabilities. Moreover, the conditional independent assumption is also common in identifying temporal latent processes \cite{kong2022partial,li2023subspace}. Intuitively, it means there are no immediate relations among latent variables. To satisfy this assumption, we can sample data at high frequency to avoid instantaneous dependencies caused by subsampling. 

\textbf{Sufficient Variability.} The implications of the sufficient variability of historical information or environment in Theorem \ref{the:block} are similar. It is also common in \cite{yao2022temporally,kong2022partial}, reflecting that the influence of each latent variable on the observations is independent. This assumption is also a standard requirement for identifying nonlinear ICA \cite{allman2009identifiability,hyvarinen2016unsupervised,halva2020hidden,lippe2022citris}, ensuring a unique solution to the system of equations. Although this assumption is untestable, it can be assessed based on prior knowledge specific to the application.

\textbf{Prior Environment Number.} The prior environment number assumption implies that we can take the number of environments as prior knowledge. For example, we can know the number of actions of drivers.

\textbf{Full Rank and Linear Linear Independence.} The full-rank assumption implies that the state transition matrix of latent environment is full-rank, meaning the transition probability between any two environments is nonzero. However, if the collected data is insufficient, it may not capture all environment transitions, causing this assumption to break down. To meet this assumption, we should collect as much data as possible across diverse environmental conditions. 

\textbf{Linear Independent.} The Linear Independence assumption aligns closely with the concept of sufficient variability. It implies that when the environment changes, the resulting variations in observed variables are significant. For example, the effects of speeding and braking on a car’s speed are fundamentally different, reflecting distinct influences from the underlying environment.


\section{Identifiable Latent States Model}

\begin{figure}[t]
    \centering
\includegraphics[width=\columnwidth]{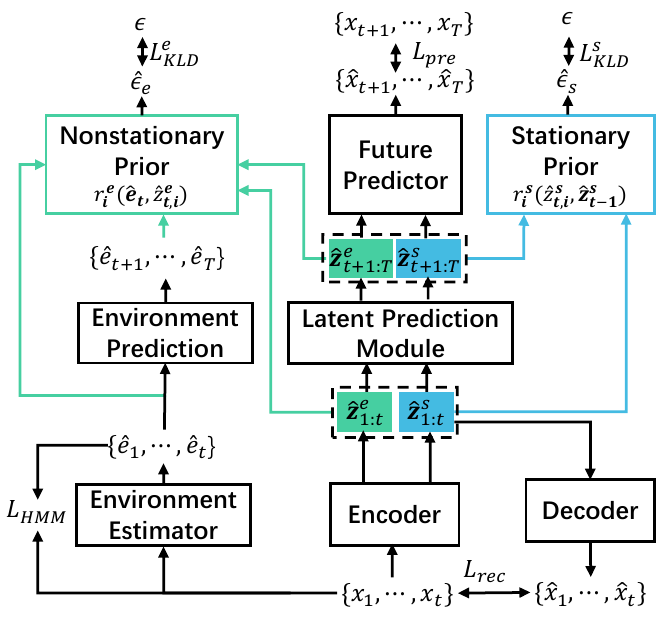} 
 \vspace{-0.5cm}
	\caption{The framework of the UDA model. The latent variable encoder is used to extract $\rvz^s_{1:t}$ and $\rvz^e_{1:t}$ from $\rvx_{1:t}$. The latent forecasting module is used to estimate $\rvz^s_{t+1:T}$ and $\rvz^e_{t+1:T}$ from $\rvz^s_{1:t}$ and $\rvz^e_{1:t}$. The future forecasting module is used for future prediction $\{\hat{x}_{t+1},\cdots,\hat{x}_T\}$. The historical latent environments $\{\hat{\rve}_1,\cdots,\hat{\rve}_{t}\}$ are generated by the environment estimation module, and the future latent environments $\{\hat{\rve}_{t+1},\cdots,\hat{\rve}_{T}\}$ are generated by the environment prediction module. The nonstationary prior and the stationary prior are used to estimate the prior distribution of stationary and nonstationary latent variables for KL divergence.}  
 \label{fig:model}
 \vspace{-0.25cm}
\end{figure}

In this section, we introduce the implementation of the UDA model as shown in Figure \ref{fig:model}, which is built on a sequential variational inference module with an autoregressive hidden Markov Module for latent environment estimation. Moreover, we devise modular prior networks to estimate the prior of stationary and nonstationary latent variables.

\subsection{Sequential Variational Inference Module for Time Series Data Modeling}
Based on the data generation process in Figure \ref{fig:data_generation1}, we first derive the evidence lower bound (ELBO) in Equation (\ref{equ:elbo}). Please refer to Appendix \ref{app:elbo} for more details of the derivation.  
\begin{equation}
\label{equ:elbo}
\small
\begin{split}
ELBO=&\mathcal{L}_{pre} \!+\! \alpha \mathbb{E}_{q(\rvz^e_{1:t}|\rvx_{1:t}\!)}\mathbb{E}_{q(\rvz^s_{1:t}|\rvx_{1:t}\!)}\mathcal{L}_{rec} \\& - \beta \mathcal{L}_{KLD}^s \!-\! \gamma\mathcal{L}_{KLD}^e
\end{split}
\end{equation}
\vspace{-1mm}
where $\alpha, \beta$ and $\gamma$ denote the hyper-parameters. Note that $\mathcal{L}_{rec}$ and $\mathcal{L}_{pre}$ denote the reconstruction of historical observations and future predictor module shown as follows:
\begin{equation}
\small
\begin{split}
\mathcal{L}_{rec}\!=&\mathbb{E}_{q(\rvz^e_{1:t}|\rvx_{1:t})}\mathbb{E}_{q(\rvz^s_{1:t}|\rvx_{1:t})}\ln p(\rvx_{1:t}|\rvz^e_{1:t},\rvz^s_{1:t})\\
\mathcal{L}_{pre}\!=&\mathbb{E}_{q(\rvz^e_{t:T}|\rvz^e_{1:t})}\mathbb{E}_{q(\rvz^s_{t:T}|\rvz^s_{1:t})}\ln p(\rvx_{t+1:T}|\rvz^e_{t+1:T},\rvz^s_{t+1:T}).
\end{split}
\end{equation}
$\mathcal{L}_{KLD}^s$ and $\mathcal{L}_{KLD}^e$ denote the Kullback-Leibler divergence between the approximated posterior distribution and the estimated prior distribution as shown in Equation (\ref{equ:KL}):
\begin{equation}
\small
\begin{split}
\mathcal{L}_{KLD}^s\!&=D_{KL}\!(q(\rvz^s_{1:t}|\rvx_{1:t})||p(\rvz^s_{1:t}))\\&+\mathbb{E}_{q(\rvz^s_{1:t}|\rvx_{1:t})}\!\Big[\!D_{KL}\!(q(\rvz^s_{t+1:T}|\rvz^s_{1:t})||p(\rvz^s_{t+1:T}|\rvz^s_{1:t}))\Big]\\
\mathcal{L}_{KLD}^e\!&=D_{KL}\!(q(\rvz^e_{1:t}|\rvx_{1:t})||p(\rvz^e_{1:t}))\\&+\mathbb{E}_{q(\rvz^e_{1:t}|\rvx_{1:t})}\!\Big[\!D_{KL}\!(q(\rvz^e_{t+1:T}|\rvz^e_{1:t})||p(\rvz^e_{t+1:T}|\rvz^e_{1:t}))\Big],
\end{split}
\label{equ:KL}
\end{equation}
in which $q(\rvz_{1:t}^s|\rvx_{1:t}),\, \, q(\rvz_{t+1:T}^s|\rvz_{1:t}^s), \, \, q(\rvz_{1:t}^e|\rvx_{1:t})$ and $q(\rvz_{t+1:T}^e|\rvz_{1:t}^e)$ are used to approximate the distribution. Therefore, the aforementioned approximate functions, the historical decoder, and the future forecasting module can be formalized as follows:
\begin{gather}
\small
\label{equ:model}
\begin{split}
\hat\rvz^e_{1:t}&=\psi_e(\rvx_{1:t};\theta_{\psi_e}), \quad\quad\,\,\,\,\quad
\hat\rvz^s_{1:t}=\psi_s(\rvx_{1:t};\theta_{\psi_s}),\,\,\,\,\quad
\\\hat\rvz^e_{t+1:T}&=T_e(\hat\rvz^e_{1:t};\theta_{T_e}),\,\,\,\,\quad \quad \hat\rvz^s_{t+1:T}=T_s(\hat\rvz^s_{1:t};\theta_{T_s}), \,\,\,\,\quad
\\\hat{\rvx}_{1:t}&=F_x(\hat\rvz^e_{1:t},\hat\rvz^s_{1:t};\theta_x),\quad
\hat{\rvx}_{t+1:T}=F_y(\hat\rvz^e_{t+1:T},\hat\rvz^s_{t+1:T};\theta_y),
\end{split}
\end{gather}
where $\psi_s,\psi_e$ denote the latent variable encoder of stationary and nonstationary latent variables; $T_s, T_e$ are latent prediction modules; and $F_x, F_y$ denote the decoder of historical observations and the future forecasting module, respectively, which are all implemented by Multi-layer Perceptron networks (MLPs); and $\theta_{\psi_e}, \theta_{\psi_s}, \theta_{T_e},\theta_{T_s}$, and $\theta_x, \theta_y$ are the trainable parameters of neural networks.

$p(\rvz^s_{1:t}), \,\,p(\rvz^s_{t+1:T}|\rvz^s_{1:t}),\,\,p(\rvz^e_{1:t})$ and $p(\rvz^e_{t+1:T}|\rvz^e_{1:t})$ are the estimated prior distribution of stationary and nonstationary latent variables, which will be introduced in the Subsection \ref{sec:stationary_prior}. Note that the environment estimation module does not appear in the ELBO explicitly, which is a part of the $p(\rvz^e_{1:t}), p(\rvz^e_{t+1:T}|\rvz^e_{1:t})$ and will be also introduced in subsection \ref{sec:stationary_prior}. Please refer to the Appendix \ref{app:implementation} for more details of the implementation of the proposed \textbf{UDA} model.

\subsection{Stationary and Nonstationary Priors Estimation}\label{sec:stationary_prior}

Previous time-series modeling methods based on the causality-based data generation processes usually require autoregressive inference and Gaussian prior \cite{fabius2014variational,kim2019variational,zhu2020s3vae}. However, simply assuming the Gaussian distribution might result in suboptimal performance of disentanglement. To solve this problem, we employ the modular neural architecture to estimate the prior distribution of stationary and nonstationary latent variables.


\textbf{Modular Architecture for the Stationary Prior Estimation}. we first let $\{r_i^s\}$ be a set of learned inverse transition functions that take the estimated stationary latent variables and output the noise term, i.e., $\hat{\epsilon}_{t,i}^s=r_i^s(\hat{z}_{t,i}^s, \hat{\rvz}^s_{t-1})$ \footnote{We use the superscript symbol to denote estimated variables.} and each $r_i^s$ is modeled with MLPs. Then we devise a transformation $\phi^s:=\{\hat{\rvz}^s_{t-1},\hat{\rvz}^s_t\}\rightarrow\{\hat{\rvz}^s_{t-1}, \hat{\epsilon}^s_t\}$, and its Jacobian is $\small{\mathbf{J}_{\phi^s}=
    \begin{pmatrix}
        \mathbb{I}&0\\
        * & \text{diag}\left(\frac{\partial r^s_i}{\partial \hat{z}_{t,i}^s}\right)
    \end{pmatrix}}$,
where $*$ denotes a matrix. By applying the change of variables formula, we have:
\begin{equation}
\small
    \log p(\hat{\rvz}^s_{t-1},\hat{\rvz}^s_t)=\log p(\hat{\rvz}^s_{t-1},\hat{\epsilon}_t^s) + \log|\text{det}(\mathbf{J}_{\phi})|.
\end{equation}
Since we assume that the noise term in Equation (\ref{equ:data_gen2}) is independent of $\rvz_{t-1}^s$, we enforce the independence of the estimated noise $\hat{\epsilon}_t^s$ and we further have:
\begin{equation}
\small
\log p(\hat{\rvz}^s_{t}|\hat{\rvz}^s_{t-1})=\log p(\hat{\epsilon}^s_{t}) + \sum_{i=n_d+1}^n\log|\frac{\partial r_i^s}{\partial \hat{z}^s_{t,i}}|.
\end{equation}
Therefore, the stationary prior can be estimated as follows:
\begin{equation}
\small
p(\hat{\rvz}_{1:t}^s)=p(\hat{\rvz}^s_1)+\prod_{\tau=2}^t\!\! \left( \sum_{i=n_d+1}^n\log p(\hat{\epsilon}^s_{\tau,i}) +\!\!\!\!\sum_{i=n_d+1}^n\!\!\!\log|\frac{\partial r_i^s}{\partial \hat{z}^s_{\tau,i}}| \right),
\end{equation}
where $p(\hat{\epsilon}^s_i)$ follow Gaussian distributions. Another prior $p(\hat{\rvz}_{t+1:T}^s|\hat{\rvz}_{1:t}^s)$ follows a similar derivation.

\textbf{Modular Architecture for the Nonstationary Prior Estimation}. We employ a similar derivation and let $\{r_i^e\}$ be a set of learned inverse transition functions, which take the estimated environment labels $\hat{\rve}_t$ and $\hat{\rvz}^e_t$ as input and output the noise term, i.e. $\hat{\epsilon}^e_t=r_i^e(\hat{\rve}_t, \hat{z}^e_{t,i})$. Leaving $r_i^e$ be an MLP, we further devise another transformation $\phi^e:=\{\hat{\rve}_t,\hat{\rvz}^e_t\} \rightarrow \{\hat{\rve}_t,\hat{\epsilon}^e_t\}$ with its Jacobian $\small{\mathbf{J}_{\phi^e}=
    \begin{pmatrix}
        \mathbb{I}&0\\
        * & \text{diag}\left(\frac{\partial r_i^e}{\partial \hat{z}_{t,i}^e}\right)
\end{pmatrix}}$, where $*$ denotes a matrix. Similarly, we have:
\begin{equation}
\small
\ln p(\hat{\rvz}_t^e|\hat{\rve}_t)=\ln p(\hat{\epsilon}^e_t) + \sum_{i=1}^{n_e}\ln |\frac{\partial r_i^e}{\partial \hat{z}^e_{t,i}}|.
\end{equation}
Therefore, the nonstationary prior can be estimated by maximizing the following equation:
\begin{equation}
\label{equ:nonstationary_prior}
\small
\begin{split}
\ln p(\hat{\rvz}^e_{1:t})&=\mathbb{E}_{q(\hat{\rve}_{1:t})}\sum_{\tau=1}^t\left(\sum_{i=1}^{n_e}\ln p(\hat{\epsilon}^e_{\tau,i}) + \sum_{i=1}^{n_e}\ln |\frac{\partial r_i^e}{\partial \hat{z}^e_{\tau,i}}|\right).
\end{split}
\end{equation}
Note that $q(\hat{\rve}_{1:t})$ denotes the environment estimation module, which is implemented with an autoregressive hidden Markov model and generates latent environment indices with the help of the Viterbi Algorithm \cite{song2023temporally,elliott2012viterbi}. To optimize the autoregressive hidden Markov model, we need to maximize its free energy lower bound, which is shown as follows:
\begin{equation}
\small
\begin{split}
\ln p(\rvx_{1:t})=&\mathbb{E}_{q(\rve_{1:t})}\ln\frac{p(\rvx_{1:t},\rve_{1:t})q(\rve_{1:t})}{p(\rve_{1:t}|\rvx_{1:t})q(\rve_{1:t})}\\\geq&\mathbb{E}_{q(\rve_{1:t})}p(\rve_{1:t}|\rvx_{1:t}) - \textbf{H}(q(\rve_{1:t}))=\mathcal{L}_{HMM}.
\end{split}
\end{equation}
Please refer to more detailed derivations of stationary and nonstationary in Appendix \ref{app:prior}.

\subsection{Model Summary}
Considering that the autoregressive hidden Markov model converges much faster than the sequential variational inference model, we employ a two-phase training strategy. Specifically, we first minimize $\mathcal{L}_{HMM}$ to train the autoregressive hidden Markov model. Then we minimize $\mathcal{L}_{ELBO}$ by fixing the parameters of the autoregressive hidden Markov model.
Since we use the historical observations $\rvx_{1:t}$ to generate $\hat{\rve}_{1:t}$, which can be used to estimate the transition matrix $\hat{\mathbf{A}}$, during testing phase, we can estimate $\hat{\rve}_{t+1:T}$ by sampling from $\hat{\mathbf{A}}$ as shown in the environment prediction block in Figure \ref{fig:model}. Please refer to Appendix \ref{app:eff} for model efficiency comparison.

\section{Experiments}
\subsection{Synthetic Experiments}

\subsubsection{Experimental Setup} \label{sim_exp}
\textbf{Data Generation.} We generate the simulated nonstationary time series data with 3 environments. 
Specifically, we first randomly initialize a Markov Chain with a transition matrix $\textbf{A}$. Sequentially, for each environment, we consider different Gaussian distributions and generate the nonstationary latent variables $\rvz_t^e$. As for the stationary latent variables, we employ an MLP with a LeakyReLU unit as the transition function. We generate Dataset A and Dataset B with different time lag dependencies. Finally, we use a randomly initialized MLP to generate the observation data.

\textbf{Evaluation Metrics.} We consider three different metrics to evaluate the effectiveness of our method. First, to evaluate the identifiability of the stationary and nonstationary latent variables, we consider the Mean Correlation Coefficient (MCC) on the test dataset, which is a standard metric for nonlinear ICA. A higher MCC denotes that the model can achieve better identification performance. Second, to evaluate the identifiability of the transition matrix \textbf{A}, we further consider the Mean Square Error (MSE) between the ground truth \textbf{A} and the estimated one. A lower value of MSE implies the model can identify the transition matrix better. Finally, we also consider the accuracy of $\rve_t$ estimation since it reflects the performance of our model in detecting when the temporal distribution shift occurs. Please refer to Appendix \ref{app:simulation} for a detailed discussion about evaluation metrics.

\textbf{Baselines.} Besides the standard BetaVAE \cite{higgins2016beta} that does not consider any temporal and environment information, we also take some conventional nonlinear ICA methods into account like TCA \cite{hyvarinen2016unsupervised}and i-VAE \cite{khemakhem2020variational}. Moreover, we consider TDRL \cite{yao2022temporally}, which considers stationary and nonstationary causal dynamics, but requires observed environment variables. Finally, we consider the HMNLICL \cite{halva2020hidden}, NCTRL \cite{song2023temporally}, and CtrlNS \cite{song2024causal}, which are designed for unobserved nonstationary.

\subsubsection{Results and Discussion}

\begin{table}
\centering
\caption{Experiment results of two synthetic datasets on baselines and proposed \textbf{UDA}.}
\resizebox{\columnwidth}{!}{%
\begin{tabular}{@{}c|ccc@{}}
\toprule
\multirow{2}{*}{Method} & \multicolumn{3}{c}{Mean Correlation Coefficien (MCC)} \\ \cmidrule(l){2-4} 
                        & Dataset A        & Dataset B        & Average         \\ \midrule
\small{\textbf{BetaVAE} \cite{higgins2016beta}}                & 64.2             & 63.2             & 63.7            \\ \small{\textbf{TCL} \cite{hyvarinen2016unsupervised}}                    & 56.0               & 65.8             & 60.9            \\
\small{\textbf{i-VAE} \cite{khemakhem2020variational} }                 & 76.9             & 73.0               & 74.9           \\
\small{\textbf{HMNLICA} \cite{halva2020hidden}}                & 83.2             & 74.5             & 78.8           \\
\small{\textbf{TDRL} \cite{yao2022temporally}}                   & 78.5             & 78.8             & 78.6           \\
\small{\textbf{NCTRL} \cite{song2023temporally}}                   & 81.4             & 79.4             & 80.4            \\ 
\small{\textbf{CtrlNS} \cite{song2024causal}}                   & 87.9             & 85.1             & 86.5            \\ 
\midrule
\small{\textbf{UDA}}                    & \textbf{97.5}    & \textbf{92.7}    & \textbf{95.1}   \\ \bottomrule
\end{tabular}%
}
\label{tab:simulation1}
\vspace{-3mm}
\end{table}
Experiment results of MCC are shown in Table \ref{tab:simulation1}, and the experiment results of environment estimation accuracy and MSE can be found in Appendix \ref{app:simulation}. We can draw the following conclusions: 1) We can find that the accuracies of environment estimation are high in both datasets. Since Dataset B contains more complex temporal relationships, the corresponding accuracy is slightly lower. 2) We can also find that the proposed \textbf{UDA} model can reconstruct the latent variables under unknown temporal distribution shift with ideal MCC performance, i.e., $(>0.95)$ on average. In the meanwhile, the other compared methods, which do not use historical dependency, can hardly perform well. Moreover, the TDRL, which considers the temporal causal relationship, cannot obtain an ideal MCC performance since it requires observed environments. 3) Although baselines like HMNLICA, NCTRL and CtrlNS are devised for unobserved nonstationarity, they require strong conditions and hence can not achieve ideal identification results.

\subsection{Real-world Experiments}

\subsubsection{Experiment Setup}
\textbf{Datasets. }We conduct experiments on eight real-world benchmark datasets that are widely used in nonstationary time series forecasting: ETT \cite{zhou2021informer}, Exchange \cite{lai2018modeling}, ILI(CDC), electricity consuming load (ECL), weather (Wetterstation), traffic and M4 \cite{makridakis2020m4}. More detailed descriptions of the datasets can be found in  Appendix \ref{app:dataset}. We employ the same data preprocessing and split ratio in TimeNet \cite{wu2022timesnet}. Following the same setting of TimesNet, for each forecasting window length $H$, we let the length of the lookback window be $H$. Moreover, for each dataset, we consider different forecast lengths $H\in \{48, 96, 144, 192\}$.

\textbf{Baselines.} We consider the following state-of-the-art deep forecasting models for time series data. First, we consider the methods for long-term forecasting including the TCN-based methods like TimesNet \cite{wu2022timesnet} and  MICN \cite{wang2022micn}, and ModernTCN \cite{luo2024moderntcn} the MLP-based methods like DLinear \cite{zeng2023transformers}, as well as the recently proposed WITRAN \cite{jia2023witran}, iTransformer \cite{liu2023itransformer}, and FITS \cite{xu2023fits}. Moreover, we further consider the methods with the assumption that the temporal distribution shift occurs among instances like RevIN \cite{kim2021reversible} and Nonstationary Transformer \cite{liu2022non}. Finally, we compare the nonstationary forecasting methods with the assumption that temporal distribution shift occurs uniformly, like Koopa \cite{liu2023koopa} and SAN \cite{liu2023adaptive}.
We also consider the recent works for nonstationary time series forecasting like FOIL \cite{liu2024time}, SOILD \cite{liu2024timebridge}, and FAN \cite{ye2024frequency}. We repeat each experiment over 3 random seeds and publish the average performance. Please refer to Appendix \ref{app:other_exp} for more experiments.
\begin{table*}[]
\renewcommand{\arraystretch}{1}
\caption{MSE and MAE results on the ETTh1, ETTh2, Exchange, ILI, Weather, Traffic, and ECL datasets. N-Transformer denotes the nonstationary Transformer due to the limited space.}
\centering
\label{tab:main_exp}
\resizebox{\textwidth}{!}{%
\begin{tabular}{cccccccccccccccccccc}
\hline
\multicolumn{2}{c}{Models}                                                  & \multicolumn{2}{c}{UDA}        & \multicolumn{2}{c}{Koopa}       & \multicolumn{2}{c}{iTransformer} & \multicolumn{2}{c}{Informer+FOIL} & \multicolumn{2}{c}{PatchTST+SOILD} & \multicolumn{2}{c}{DLinear+FAN} & \multicolumn{2}{c}{TimesNet} & \multicolumn{2}{c}{DLinear} & \multicolumn{2}{c}{N-Transformer} \\ \hline
\multicolumn{1}{c|}{Metric}                   & \multicolumn{1}{c|}{Length} & MSE            & MAE            & MSE            & MAE            & MSE                 & MAE        & MSE             & MAE             & MSE          & MAE                 & MSE            & MAE            & MSE           & MAE          & MSE      & MAE              & MSE             & MAE             \\ \hline
\multicolumn{1}{c|}{\multirow{4}{*}{ECL}}     & \multicolumn{1}{c|}{48}     & \textbf{0.129} & \textbf{0.228} & 0.13           & 0.234          & 0.15                & 0.242      & 0.219           & 0.314           & 0.179        & 0.262               & 0.195          & 0.275          & 0.149         & 0.254        & 0.158    & 0.241            & 0.155           & 0.26            \\
\multicolumn{1}{c|}{}                         & \multicolumn{1}{c|}{96}     & \textbf{0.131} & \textbf{0.223} & 0.136          & 0.236          & 0.138               & 0.233      & 0.231           & 0.322           & 0.185        & 0.267               & 0.194          & 0.278          & 0.17          & 0.275        & 0.153    & 0.245            & 0.175           & 0.279           \\
\multicolumn{1}{c|}{}                         & \multicolumn{1}{c|}{144}    & \textbf{0.147} & \textbf{0.239} & 0.149          & 0.247          & \textbf{0.146}      & 0.241      & 0.246           & 0.337           & 0.182        & 0.266               & 0.189          & 0.275          & 0.183         & 0.287        & 0.152    & 0.245            & 0.189           & 0.289           \\
\multicolumn{1}{c|}{}                         & \multicolumn{1}{c|}{192}    & 0.157          & 0.25           & 0.156          & 0.254          & \textbf{0.152}      & 0.248      & 0.28            & 0.363           & 0.189        & 0.272               & 0.192          & 0.281          & 0.189         & 0.291        & 0.153    & \textbf{0.246}   & 0.197           & 0.298           \\ \hline
\multicolumn{1}{c|}{\multirow{4}{*}{ETTh2}}   & \multicolumn{1}{c|}{48}     & \textbf{0.225} & \textbf{0.298} & 0.226          & 0.3            & 0.244               & 0.314      & 0.258           & 0.407           & 0.248        & 0.32                & 0.344          & 0.373          & 0.241         & 0.319        & 0.226    & 0.305            & 0.318           & 0.375           \\
\multicolumn{1}{c|}{}                         & \multicolumn{1}{c|}{96}     & \textbf{0.284} & \textbf{0.34}  & 0.297          & 0.349          & 0.302               & 0.356      & 0.302           & 0.369           & 0.313        & 0.359               & 0.386          & 0.399          & 0.325         & 0.376        & 0.294    & 0.351            & 0.411           & 0.441           \\
\multicolumn{1}{c|}{}                         & \multicolumn{1}{c|}{144}    & \textbf{0.312} & \textbf{0.365} & 0.333          & 0.381          & 0.353               & 0.389      & 0.335           & 0.456           & 0.358        & 0.385               & 0.421          & 0.424          & 0.374         & 0.408        & 0.353    & 0.397            & 0.48            & 0.469           \\
\multicolumn{1}{c|}{}                         & \multicolumn{1}{c|}{192}    & \textbf{0.336} & \textbf{0.379} & 0.356          & 0.393          & 0.383               & 0.408      & 0.499           & 0.482           & 0.397        & 0.408               & 0.445          & 0.443          & 0.394         & 0.434        & 0.385    & 0.418            & 0.449           & 0.467           \\ \hline
\multicolumn{1}{c|}{\multirow{4}{*}{Exhange}} & \multicolumn{1}{c|}{48}     & \textbf{0.042} & 0.141          & \textbf{0.042} & 0.143          & 0.045               & 0.148      & 0.063           & 0.146           & 0.0426       & \textbf{0.083}      & 0.042          & 0.142          & 0.059         & 0.172        & 0.043    & 0.145            & 0.07            & 0.188           \\
\multicolumn{1}{c|}{}                         & \multicolumn{1}{c|}{96}     & 0.086          & \textbf{0.205} & \textbf{0.083} & 0.207          & 0.094               & 0.219      & 0.142           & 0.274           & 0.09         & 0.222               & 0.088          & 0.21           & 0.12          & 0.255        & 0.084    & 0.22             & 0.171           & 0.296           \\
\multicolumn{1}{c|}{}                         & \multicolumn{1}{c|}{144}    & \textbf{0.125} & \textbf{0.254} & 0.13           & 0.261          & 0.157               & 0.284      & 0.184           & 0.341           & 0.137        & 0.298               & 0.128          & 0.258          & 0.206         & 0.334        & 0.132    & 0.254            & 0.35            & 0.416           \\
\multicolumn{1}{c|}{}                         & \multicolumn{1}{c|}{192}    & \textbf{0.164} & \textbf{0.296} & 0.184          & 0.309          & 0.214               & 0.336      & 0.236           & 0.369           & 0.177        & 0.331               & 0.171          & 0.297          & 0.377         & 0.463        & 0.178    & 0.299            & 0.566           & 0.573           \\ \hline
\multicolumn{1}{c|}{\multirow{4}{*}{ILI}}     & \multicolumn{1}{c|}{24}     & \textbf{1.456} & \textbf{0.778} & 1.621          & 0.800          & 2.422               & 1.018      & 2.184           & 0.806           & 1.477        & 0.858               & 1.556          & 0.784          & 2.464         & 1.039        & 2.624    & 1.118            & 2.565           & 1.018           \\
\multicolumn{1}{c|}{}                         & \multicolumn{1}{c|}{36}     & \textbf{1.79}  & \textbf{0.839} & 1.803          & 0.855          & 2.491               & 1.050      & 2.956           & 1.121           & 1.873        & 0.909               & 1.832          & 0.859          & 2.388         & 1.007        & 2.693    & 1.156            & 1.997           & 0.951           \\
\multicolumn{1}{c|}{}                         & \multicolumn{1}{c|}{48}     & \textbf{1.746} & \textbf{0.885} & 1.768          & 0.903          & 2.353               & 1.049      & 2.570           & 1.188           & 1.914        & 1.038               & 1.788          & 0.899          & 2.370         & 1.040        & 2.852    & 1.229            & 2.165           & 0.988           \\
\multicolumn{1}{c|}{}                         & \multicolumn{1}{c|}{60}     & 1.831          & \textbf{0.89}  & \textbf{1.743} & 0.891          & 2.542               & 1.105      & 2.635           & 1.109           & 1.976        & 0.995               & 1.909          & 0.893          & 2.193         & 1.003        & 2.554    & 1.144            & 2.163           & 1.049           \\ \hline
\multicolumn{1}{c|}{\multirow{4}{*}{Traffic}} & \multicolumn{1}{c|}{48}     & \textbf{0.36}  & \textbf{0.231} & 0.415          & 0.274          & 0.372               & 0.249      & 0.522           & 0.29            & 0.432        & 0.354               & 0.532          & 0.414          & 0.567         & 0.306        & 0.488    & 0.352            & 0.541           & 0.35            \\
\multicolumn{1}{c|}{}                         & \multicolumn{1}{c|}{96}     & \textbf{0.323} & \textbf{0.218} & 0.401          & 0.275          & 0.337               & 0.233      & 0.53            & 0.293           & 0.401        & 0.325               & 0.476          & 0.375          & 0.611         & 0.337        & 0.485    & 0.336            & 0.529           & 0.349           \\
\multicolumn{1}{c|}{}                         & \multicolumn{1}{c|}{144}    & \textbf{0.315} & \textbf{0.217} & 0.397          & 0.276          & 0.323               & 0.229      & 0.558           & 0.305           & 0.389        & 0.313               & 0.431          & 0.346          & 0.603         & 0.322        & 0.452    & 0.317            & 0.538           & 0.353           \\
\multicolumn{1}{c|}{}                         & \multicolumn{1}{c|}{192}    & \textbf{0.315} & \textbf{0.22}  & 0.403          & 0.284          & 0.322               & 0.233      & 0.589           & 0.328           & 0.399        & 0.308               & 0.434          & 0.348          & 0.604         & 0.321        & 0.438    & 0.309            & 0.507           & 0.342           \\ \hline
\multicolumn{1}{c|}{\multirow{4}{*}{Weather}} & \multicolumn{1}{c|}{48}     & \textbf{0.124} & \textbf{0.167} & 0.126          & 0.168          & 0.14                & 0.179      & 0.177           & 0.218           & 0.148        & 0.188               & 0.158          & 0.217          & 0.138         & 0.191        & 0.156    & 0.198            & 0.143           & 0.195           \\
\multicolumn{1}{c|}{}                         & \multicolumn{1}{c|}{96}     & \textbf{0.151} & 0.301          & 0.154          & \textbf{0.205} & 0.168               & 0.214      & 0.225           & 0.259           & 0.187        & 0.226               & 0.199          & 0.265          & 0.18          & 0.231        & 0.186    & 0.229            & 0.199           & 0.246           \\
\multicolumn{1}{c|}{}                         & \multicolumn{1}{c|}{144}    & 0.177          & \textbf{0.22}  & \textbf{0.172} & 0.225          & 0.184               & 0.232      & 0.278           & 0.297           & 0.207        & 0.242               & 0.213          & 0.274          & 0.19          & 0.244        & 0.199    & 0.244            & 0.225           & 0.267           \\
\multicolumn{1}{c|}{}                         & \multicolumn{1}{c|}{192}    & \textbf{0.193} & \textbf{0.233} & 0.193          & 0.241          & 0.203               & 0.252      & 0.354           & 0.348           & 0.234        & 0.265               & 0.238          & 0.298          & 0.212         & 0.265        & 0.217    & 0.261            & 0.296           & 0.315           \\ \hline
\end{tabular}%
}
\end{table*}

\subsubsection{Results and Discussion}


\textbf{Quantitative
Results.} Experiment results on each dataset are shown in Table \ref{tab:main_exp}. Please refer to Appendix \ref{app:other_exp} and \ref{app:sensitive} for experiment results on other datasets and sensitivity analysis. Based on the experimental results, our UDA model significantly outperforms all other baselines in most forecasting tasks. Specifically, it exceeds the performance of the most competitive baselines by a clear margin of $1.7\%–10\%$, and substantially reduces forecasting errors on challenging benchmarks such as weather and ILI. In addition to outperforming forecasting models that do not account for nonstationary assumptions, like TimesNet and DLinear, our UDA model also excels with nonstationary time series data, such as nonstationary Transformer. However, in the Exchange dataset with a forecasting length of 72, our method achieves the second-best results, still comparable to the top performer. This may be attributed to inaccuracies in environmental estimation for long-term predictions.

It is remarkable that our method achieves a better performance than that of FOIL and SOILD, which assume temporal distribution shifts in each time series instance. This is because these methods assume that uniform temporal distribution shifts in each time series instance, which is hard to meet in real-world scenarios, and it is hard for these methods to disentangle the stationary and nonstationary components simultaneously. Meanwhile, our method detects when the temporal distribution shift occurs and further disentangles the stationary and nonstationary states with identification guarantees, hence it can achieve the ideal nonstationary forecasting performance. Please refer to Appendix \ref{app:other_exp} for experiment results on the M4 dataset.


\subsubsection{Ablation Study}
We further devise three model variants. a) \textbf{UDA}-H: We remove the autoregressive hidden Markov model for environment estimation, and use random environment variables. b) \textbf{UDA}-E: We remove the nonstationary prior and the corresponding KL divergence term. c) \textbf{UDA}-S: We remove the stationary prior and the corresponding Kullback-Leibler divergence term. d) \textbf{UDA}-sh: We use a shared decoder for forecasting and reconstruction. Experiment results on the ILI dataset are shown in Figure \ref{fig:ablation} in Appendix \ref{app:ablation}. We find that 1) the performance of \textbf{UDA}-H drops without an accurate estimation of the environments, implying that the accurate environment estimation benefits the disentanglement and forecasting performance. 2) Both stationary and non-stationary priors play an important role in forecasting, implying that these priors can capture temporal information. 

\section{Conclusion}

This paper introduces the UDA for nonstationary time series forecasting that addresses the challenge of adapting to temporal distribution shifts without relying on the assumption of uniformity. By detecting when distribution shifts occur and disentangling stationary and nonstationary latent variables, the UDA model enables dynamic adaptation to evolving environments. We leverage a Hidden Markov model to identify latent environments and use the variability of historical data to effectively separate stationary and nonstationary components. Through the integration of variational autoencoders and modular prior networks, our model facilitates automatic adaptation to nonstationary changes, significantly enhancing forecasting performance. Experimental results on multiple datasets demonstrate the practical effectiveness and superiority of our approach, marking a key advancement in the field of nonstationary time series forecasting.







\clearpage
\section{Impact Statement}

The proposed \textbf{UDA} model can detect when the temporal distribution shifts occur and disentangle the stationary and nonstationary latent variables. Therefore, our \textbf{UDA} could be applied to a wide range of applications including time series forecasting, imputation, and classification. Specifically, the disentangled stationary and nonstationary latent variables would create a model that is more transparent, thereby aiding in the reduction of bias and the promotion of fairness of the existing time series forecasting models. 
\nocite{langley00}
\bibliography{main}
\bibliographystyle{apalike}

\newpage
\appendix
\onecolumn

\section{Related Works}\label{app:related_works}
We review the works about nonstationary time series forecasting and the identifiability of latent variables.

\textbf{Nonstationary Time Series Forecasting.} 
Time series forecasting is a conventional task in the field of machine learning with lots of successful cases, e.g, autoregressive model \cite{hyndman2018forecasting} and ARMA \cite{box1970distribution}. Previously, deep neural networks also have made great contributions to time series forecasting, e.g., RNN-based models \cite{hochreiter1997long,lai2018modeling,salinas2020deepar}, CNN-based models \cite{bai2018empirical,wang2022micn,wu2022timesnet}, and the methods based on state-space model \cite{gu2022parameterization,gu2020hippo,gu2021combining,gu2021efficiently,smith2022simplified}. Recently, transformer-based methods \cite{zhou2021informer,wu2021autoformer,liu2023itransformer,nie2022time} have boosted the development of time series forecasting.

However, these methods are devised for stationary time series, so nonstationary forecasting is receiving more and more attention \cite{hu2024twins}. One straightforward solution to this challenge is to discard the nonstationarity via preprocessing methods like stationarization \cite{virili2000nonstationarity} and differencing \cite{salles2019nonstationary}, but they might destroy the temporal dependency. Recent studies have used two different assumptions to further solve this problem. By assuming that the temporal distribution shift occurs among datasets and each sequence instance is stationary \cite{cai2021time,eldele2023contrastive}, some methods consider normalization-based methods. Kim et al. \cite{kim2021reversible} propose the reversible instance normalization to remove and restore the statistical information of a time-series instance. Liu et al. \cite{liu2022non} propose the nonstationary Transformer, which includes the destationary attention mechanism to recover the intrinsic non-stationary information into temporal dependencies.

By assuming that the temporal distribution shift uniformly occurs in each sequence instance and so each equal-size segmentation is stationary, other methods propose to disentangle the stationary and nonstationary components. Surana et al. \cite{surana2020koopman} and Liu et al. \cite{liu2023koopa} employ the Koopman theory \cite{korda2018linear}, which transform the nonlinear system into several linear operators, to decompose the stationary and nonstationary factors. Liu et al. \cite{liu2023adaptive} use adaptive normalization and denormlization on non-overlap equally-sized slices. However, since the temporal distribution shift may occur any time, the aforementioned two assumptions are unreasonable. To solve this problem with milder assumptions, the proposed \textbf{UDA} first identifies when the distribution shift occurs and then identifies the latent states to learn how they change over time with the help of Markov assumption of latent environment and sufficient observation assumption. 



\textbf{Identifiability of Latent Variables.} Identifiability of latent variables \cite{kong2023understanding,yan2023counterfactual,kong2023identification} plays a significant role in the explanation and generalization of deep generative models, guaranteeing that causal representation learning can capture the underlying factors and describe the latent generation process  \cite{kumar2017variational,locatello2019challenging,locatello2019disentangling,scholkopf2021toward,trauble2021disentangled,zheng2022identifiability}. Several researchers employ independent component analysis (ICA) to learn causal representation with identifiability \cite{comon1994independent,hyvarinen2013independent,lee1998independent,zhang2007kernel} by assuming a linear generation process. To extend it to the nonlinear scenario, different extra assumptions about auxiliary variables or generation processes are adopted to guarantee the identifiability of latent variables \cite{zheng2022identifiability,hyvarinen1999nonlinear, hyvarinen2023identifiability,khemakhem2020ice,li2023identifying}. Previously, Aapo et al. established the identification results of nonlinear ICA by introducing auxiliary variables e.g., domain indexes, time indexes, and class labels\cite{khemakhem2020variational,hyvarinen2016unsupervised,hyvarinen2017nonlinear,hyvarinen2019nonlinear}. 

However, these methods usually assume that the latent variables are conditionally independent and follow the exponential families distributions. Recently, Zhang et al. release the exponential family restriction \cite{kong2022partial, xie2022multi} and propose the component-wise identification results for nonlinear ICA with a certain number of auxiliary variables. They further propose the subspace Identification \cite{li2023subspace} for multi-source domain adaptation, which requires fewer auxiliary variables. In the field of sequential data modeling, Yao et al. \cite{yao2021learning,yao2022temporally} recover time-delay latent dynamics and identify their relations from sequential data under the stationary environment and different distribution shifts. 
And Lippe et al. propose the (i)-CITRIS \cite{lippe2022citris, lippe2022icitris}, which use intervention target information for identifiability of scalar and multidimensional latent causal factors. 
Moreover, H{\"a}lv{\"a} et al. \cite{halva2020hidden} and Song et al. \cite{song2023temporally} utilize the Markov assumption to provide identification guarantee of time series data without extra auxiliary variables. Although Yao et al. \cite{yao2022temporally} partitioned the latent space into stationary and nonstationary parts, they require extra environment variables. 
Furthermore, although H{\"a}lv{\"a} et al. \cite{halva2020hidden} and Song et al. \cite{song2023temporally,song2024causal} provide identifiability results without extra environment variables, they can hardly disentangle the stationary and nonstationary, respectively.  

\section{Identification} \label{app:identification}
In this section, we provide the definition of different types of identificaiton.

\subsection{Componenet-wise Identification}
For each ground-truth changing latent variables $z_{t,i}$, there exists a corresponding estimated component $\hat{z}_{t,j}$ and an invertible function $h_{t,i}: \mathbb{R}\rightarrow \mathbb{R}$, such that $\hat{z}_{t,j}=h(z_{t,i})$.

\subsection{Subspace Identification}
For each ground-truth changing latent variables $z_{t,i}$, the subspace identification means that there exists $\hat{\rvz}_{t}$ and an invertible function $z_{t,i}=h_i(\hat{\rvz}_t)$, such that $z_{t,i}=h_i(\hat{\rvz}_t)$.

\subsection{Identification Up to Label Swapping}
If $\tilde{\textbf{A}}$ is a $E\times E$ transition matrix and if $\tilde{\pi}(e)$ is a stationary distribution of $\tilde{\textbf{A}}$ with $\tilde{\pi}(e)>0, \forall e\in\{1,\cdots,E\}$ and if $\tilde{M}=\{\tilde{\mu}_1,\cdots,\tilde{\mu}_j,\cdots,\tilde{\mu}_{E}\}$ are $E$ probability distributions that verify the equality of the distribution functions $\mathbb{P}^{(3)}_{\tilde{\textbf{A}},\tilde{M}}=\mathbb{P}^{(3)}_{{\textbf{A}},{M}}$, then there exist a permutation $\sigma$ of set $\{1,\cdots,E\}$ such that for all $k,l=1,\cdots,E$, we have $\tilde{A}_{k.l}=A_{\sigma(k),\sigma(l)}$ and $\tilde{\mu}_k=\mu_{\sigma(k)}$.

\section{Prior Likelihood Derivation} \label{app:prior}
In this section, we derive the prior of $p(\hat{\rvz}_{1:t}^s)$ and $p(\hat{\rvz}_{1:t}^e)$ as follows:
\begin{itemize}
    \item We first consider the prior of $\ln p(\rvz_{1:t}^s)$. We start with an illustrative example of stationary latent causal processes with two time-delay latent variables, i.e. $\rvz^s_t=[z^s_{t,1}, z^s_{t,2}]$ with maximum time lag $L=1$, i.e., $z_{t,i}^s=f_i(\rvz_{t-1}^s, \varepsilon_{t,i}^s)$ with mutually independent noises. Then we write this latent process as a transformation map $\mathbf{f}$ (note that we overload the notation $f$ for transition functions and for the transformation map):
    \begin{equation}
    \small
\begin{gathered}\nonumber
    \begin{bmatrix}
    \begin{array}{c}
        z_{t-1,1}^s \\ 
        z_{t-1,2}^s \\
        z_{t,1}^s   \\
        z_{t,2}^s
    \end{array}
    \end{bmatrix}=\mathbf{f}\left(
    \begin{bmatrix}
    \begin{array}{c}
        z_{t-1,1}^s \\ 
        z_{t-1,2}^s \\
        \varepsilon_{t,1}^s   \\
        \varepsilon_{t,2}^s
    \end{array}
    \end{bmatrix}\right).
\end{gathered}
\end{equation}
By applying the change of variables formula to the map $\mathbf{f}$, we can evaluate the joint distribution of the latent variables $p(z_{t-1,1}^s,z_{t-1,2}^s,z_{t,1}^s, z_{t,2}^s)$ as 
\begin{equation}
\small
\label{equ:p1}
    p(z_{t-1,1}^s,z_{t-1,2}^s,z_{t,1}^s, z_{t,2}^s)=\frac{p(z_{t-1,1}^s, z_{t-1,2}^s, \varepsilon_{t,1}^s, \varepsilon_{t,2}^s)}{|\text{det }\mathbf{J}_{\mathbf{f}}|},
\end{equation}
where $\mathbf{J}_{\mathbf{f}}$ is the Jacobian matrix of the map $\mathbf{f}$, which is naturally a low-triangular matrix:
\begin{equation}
\small
\begin{gathered}\nonumber
    \mathbf{J}_{\mathbf{f}}=\begin{bmatrix}
    \begin{array}{cccc}
        1 & 0 & 0 & 0 \\
        0 & 1 & 0 & 0 \\
        \frac{\partial z_{t,1}^s}{\partial z_{t-1,1}^s} & \frac{\partial z_{t,1}^s}{\partial z_{t-1,2}^s} & 
        \frac{\partial z_{t,1}^s}{\partial \varepsilon_{t,1}^s} & 0 \\
        \frac{\partial z_{t,2}^s}{\partial z_{t-1, 1}^s} &\frac{\partial z_{t,2}^s}{\partial z_{t-1,2}^s} & 0 & \frac{\partial z_{t,2}^s}{\partial \varepsilon_{t,2}^s}
    \end{array}
    \end{bmatrix}.
\end{gathered}
\end{equation}
Given that this Jacobian is triangular, we can efficiently compute its determinant as $\prod_i \frac{\partial z_{t,i}^s}{\varepsilon_{t,i}^s}$. Furthermore, because the noise terms are mutually independent, and hence $\varepsilon_{t,i}^s \perp \varepsilon_{t,j}^s$ for $j\neq i$ and $\varepsilon_{t}^s \perp \rvz_{t-1}^s$, so we can with the RHS of Equation (\ref{equ:p1}) as follows
\begin{equation}
\small
\label{equ:p2}
\begin{split}
    p(z_{t-1,1}^s, z_{t-1,2}^s, z_{t,1}^s, z_{t,2}^s)=p(z_{t-1,1}^s, z_{t-1,2}^s) \times \frac{p(\varepsilon_{t,1}^s, \varepsilon_{t,2}^s)}{|\mathbf{J}_{\mathbf{f}}|}=p(z_{t-1,1}^s, z_{t-1,2}^s) \times \frac{\prod_i p(\varepsilon_{t,i}^s)}{|\mathbf{J}_{\mathbf{f}}|}.
\end{split}
\end{equation}
Finally, we generalize this example and derive the prior likelihood below. Let $\{r_i^s\}_{i=1,2,3,\cdots}$ be a set of learned inverse transition functions that take the estimated latent causal variables, and output the noise terms, i.e., $\hat{\epsilon}_{t,i}^s=r_i^s(\hat{z}_{t,i}^s, \{ \hat{\rvz}_{t-\tau}^s\})$. Then we design a transformation $\mathbf{A}\rightarrow \mathbf{B}$ with low-triangular Jacobian as follows:
\begin{equation}
\small
\begin{gathered}
    \underbrace{[\hat{\rvz}_{t-L}^s,\cdots,{\hat{\rvz}}_{t-1}^s,{\hat{\rvz}}_{t}^s]^{\top}}_{\mathbf{A}} \text{  mapped to  } \underbrace{[{\hat{\rvz}}_{t-L}^s,\cdots,{\hat{\rvz}}_{t-1}^s,{\hat{\epsilon}}_{t,i}^s]^{\top}}_{\mathbf{B}}, \text{ with } \mathbf{J}_{\mathbf{A}\rightarrow\mathbf{B}}=
    \begin{bmatrix}
    \begin{array}{cc}
        \mathbb{I}_{n_s\times L} & 0\\
                    * & \text{diag}\left(\frac{\partial r^s_{i,j}}{\partial {\hat{z}}^s_{t,j}}\right)
    \end{array}
    \end{bmatrix}.
\end{gathered}
\end{equation}
Similar to Equation (\ref{equ:p2}), we can obtain the joint distribution of the estimated dynamics subspace as:
\begin{equation}
    \log p(\mathbf{A})=\underbrace{\log p({\hat{\rvz}}^s_{t-L},\cdots, {\hat{\rvz}}^s_{t-1}) + \sum^{n_s}_{i=1}\log p({\hat{\epsilon}}^s_{t,i})}_{\text{Because of mutually independent noise assumption}}+\log (|\text{det}(\mathbf{J}_{\mathbf{A}\rightarrow\mathbf{B}})|)
\end{equation}
Finally, we have:
\begin{equation}
\small
    \log p({\hat{\rvz}}_t^s|\{{\hat{\rvz}}_{t-\tau}^s\}_{\tau=1}^L)=\sum_{i=1}^{n_s}p({\hat{\epsilon}_{t,i}^s}) + \sum_{i=1}^{n_s}\log |\frac{\partial r^s_i}{\partial {\hat{z}}^s_{t,i}}|
\end{equation} 
Since the prior of $p(\hat{\rvz}_{t+1:T}^s|\hat{\rvz}_{1:t}^s)=\prod_{i=t+1}^{T} p(\hat{\rvz}_{i}^s|\hat{\rvz}_{i-1}^s)$ with the assumption of first-order Markov assumption, we can estimate $p(\hat{\rvz}_{t+1:T}^s|\hat{\rvz}_{1:t}^s)$ in a similar way.
\item We then consider the prior of $\ln p(\hat{\rvz}_{1:t}^e)$. Similar to the derivation of $\ln p(\hat{\rvz}_{1:t}^s)$, we let $\{r^e_i\}_{i=1,2,3,\cdots}$ be a set of learned inverse transition functions that take the estimated latent variables as input and output the noise terms, i.e. $\hat{\epsilon}_t^e=r_i^e(\hat{e}_t,\hat{z}_{t,i}^e)$. Similarly, we design a transformation $\textbf{A}\rightarrow\textbf{B}$ with low-triangular Jacobian as follows:
\begin{equation}
\small
    \underbrace{[\hat{\rve}_t, \hat{\rvz}_t^e]^{\top}}_{\textbf{A}} \quad\quad \text{mapped to} \quad\quad \underbrace{[\hat{\rve}_t, \hat{\epsilon}_t^e]^{\top}}_{\textbf{B}}, \text{with }\quad \mathbf{J}_{\textbf{A}\rightarrow\textbf{B}}=
    \begin{bmatrix}
    \begin{array}{cc}
        \mathbb{I} & 0\\
                    * & \text{diag}\left(\frac{\partial r^e_{i,j}}{\partial {\hat{z}}^e_{t,j}}\right)
    \end{array}
    \end{bmatrix}.
\end{equation}

Since the noise $\hat{\varepsilon}_t^e$ is independent of $\hat{\rve}_t$ we have 
\begin{equation}
\ln p(\hat{\rvz}_t^e|\hat{\rve}_t)=\ln p(\hat{\varepsilon}^e_t) + \sum_{i=1}^{n_e}\ln |\frac{\partial r_i^e}{\partial \hat{z}^e_{t,i}}|.
\end{equation}
\end{itemize}

\section{Evident Lower Bound}\label{app:elbo}
In this subsection, we show the evident lower bound. We first factorize the conditional distribution according to the Bayes theorem.

\begin{equation}
\tiny
\begin{split}
&\ln p(\rvx_{t+1:T},\rvx_{1:t})=\ln\frac{p(\rvx_{t+1:T},\rvz^e_{1:T},\rvz^s_{1:T},\rvx_{1:t})}{p(\rvz^e_{1:T},\rvz^s_{1:T}|\rvx_{1:t},\rvx_{t+1:T})}=\ln\frac{p(\rvx_{t+1:T},\rvz^e_{1:t},\rvz^s_{1:t},\rvz^e_{t+1:T},\rvz^s_{t+1:T},\rvx_{1:t})}{p(\rvz^e_{1:T},\rvz^s_{1:T}|\rvx_{1:T})}
\\\\=&\mathbb{E}_{q(\rvz^s_{1:t}|\rvx_{1:t})}\mathbb{E}_{q(\rvz^s_{t+1:T}|\rvz^s_{1:t})}\mathbb{E}_{q(\rvz^e_{1:t}|\rvx_{1:t})}\mathbb{E}_{q(\rvz^e_{t+1:T}|\rvz^e_{1:t})}
\ln \frac{p(\rvx_{t+1:T}|\rvz_{t+1:T}^e,\rvz_{t+1:T}^s)p(\rvx_{1:t}|\rvz_{1:t}^e,\rvz_{1:t}^s)p(\rvz^s_{1:t})p(\rvz_{1:t}^e)p(\rvz^s_{t+1:T}|\rvz^s_{1:t})p(\rvz^e_{t+1:T}|\rvz^e_{1:t})}{q(\rvz^e_{1:t}|\rvx_{1:t})q(\rvz^e_{t+1:T}|\rvz^e_{1:t})q(\rvz^s_{1:t}|\rvx_{1:t})q(\rvz_{t+1:T}^s|\rvz^s_{1:t})} \\&+ D_{KL}(q(\rvz^s_{t+1:T}|\rvz^s_{1:t})||p(\rvz^s_{t+1:T}|\rvz^s_{1:t},\rvx_{1:T},\rvz^e_{1:T}))+D_{KL}(q(\rvz^s_{1:t}|\rvx_{1:t})||p(\rvz^s_{1:t}|\rvx_{1:T},\rvz^e_{1:T})) \\&+ D_{KL}(q(\rvz^e_{1:t}|\rvx_{1:t})||p(\rvz^e_{1:t}|\rvx_{1:t})) + D_{KL}(q(\rvz^e_{t+1:T}|\rvz^e_{1:t})||p(\rvz^e_{t+1:T}|\rvx_{1:T},\rvz^e_{1:t}))
\\\\\geq&
\mathbb{E}_{q(\rvz^s_{1:t}|\rvx_{1:t})}\mathbb{E}_{q(\rvz^s_{t+1:T}|\rvz^s_{1:t})}\mathbb{E}_{q(\rvz^e_{1:t}|\rvx_{1:t})}\mathbb{E}_{q(\rvz^e_{t+1:T}|\rvz^e_{1:t})}\ln \frac{p(\rvx_{t+1:T}|\rvz_{t+1:T}^e,\rvz_{t+1:T}^s)p(\rvz^s_{1:t}|\rvx_{1:t},\rvz^e_{1:t})p(\rvz^s_{t+1:T}|\rvz^s_{1:t})p(\rvz^e_{1:t}|\rvx_{1:t})p(\rvz^e_{t+1:T}|\rvz^e_{1:t})}{q(\rvz^e_{1:t}|\rvx_{1:t})q(\rvz^e_{t+1:T}|\rvz^e_{1:t})q(\rvz^s_{1:t}|\rvx_{1:t})q(\rvz_{t+1:T}^s|\rvz^s_{1:t})}\\\\=&
\underbrace{\mathbb{E}_{q(\rvz^s_{1:t}|\rvx_{1:t})}\mathbb{E}_{q(\rvz^e_{1:t}|\rvx_{1:t})} \ln p(\rvx_{1:t}|\rvz^s_{1:t},\rvz^e_{1:t})}_{\mathcal{L}_{rec}} + 
\underbrace{\mathbb{E}_{q(\rvz^s_{1:t}|\rvx_{1:t})}\mathbb{E}_{q(\rvz^s_{t+1:T}|\rvz^s_{1:t})}\mathbb{E}_{q(\rvz^e_{1:t}|\rvx_{1:t})}\mathbb{E}_{q(\rvz^e_{t+1:T}|\rvz^e_{1:t})}\ln p(\rvx_{t+1:T}|\rvz^s_{t+1:T},\rvz^e_{t+1:T})}_{\mathcal{L}_{pre}}\\&- \underbrace{D_{KL}(q(\rvz^s_{1:t}|\rvx_{1:t})||p(\rvz^s_{1:t}))-\mathbb{E}_{q(\rvz^s_{1:t}|\rvx_{1:t})}\Big[D_{KL}(q(\rvz^s_{t:1+T}|\rvz^s_{1:t})||p(\rvz^s_{t+1:T}|\rvz^s_{1:t}))\Big]}_{\mathcal{L}_{KLD}^s}
\\&- \underbrace{D_{KL}(q(\rvz^e_{1:t}|\rvx_{1:t})||p(\rvz^e_{1:t}))-\mathbb{E}_{q(\rvz^e_{1:t}|\rvx_{1:t})}\Big[D_{KL}(q(\rvz^e_{t+1:T}|\rvz^e_{1:t})||p(\rvz^e_{t+1:T}|\rvz^e_{1:t}))\Big]}_{\mathcal{L}_{KLD}^e}
\end{split}
\end{equation}
where $p(\rvz_{1:t}^e)$ can be further formalized as follows:
\begin{equation}
\small
\ln p(\rvz_{1:t}^e)=\mathbb{E}_{q(\rve_{1:t})}\ln \frac{p(\rvz^e_{1:t}|\rve_{1:t})p(\rve_{1:t})}{p(\rve_{1:t}|\rvz^e_{1:t})}=\mathbb{E}_{q(\rve_{1:t})}\ln \frac{p(\rvz^e_{1:t}|\rve_{1:t})p(e_{1:t})q(e_{1:t})}{p(e_{1:t}|\rvz_{1:t}^e)q(e_{1:t})}\geq \mathbb{E}_{q(\rve_{1:t})} \ln p(\rvz_{1:t}^e|\rve_{1:t}) - D_{KL}(q(e_{1:t})||p(\rve_{1:t}))
\end{equation}
Since we employ a two-phase training strategy, $D_{KL}(q(e_{1:t})||p(\rve_{1:t}))$ can be considered as a small constant term after the autoregressive HMM are well trained, so $\ln p(\rvz_{1:t}^e)$ can be approximated to $\mathbb{E}_{q(\rve_{1:t})} \ln p(\rvz_{1:t}^e|\rve_{1:t})$.

\section{Identification Guarantees}

\subsection{Identification of Latent Domain Variables $\rvu_t$}\label{app:the1}

Before providing explicit proof of our identifiability result, we first give a basic lemma that proves the identifiability of the model's parameters from the joint distribution.

\begin{lemma} \underline{(Theorem 9 in \cite{allman2009identifiability})}
\label{app:lemma}
Let $\mathbb{P}$ be a mixture in the form of Equation (\ref{equ:app_lemma}), such that for every $j$, the measures $\mu_{i,j}$ are linearly independent. Then, if $c\geq 3$, $\{\pi_i, \mu_{i,j}\}$ are identifiable from $\mathbb{P}$ up to label swapping.
\begin{equation}
\label{equ:app_lemma}
    \small
\mathbb{P}=\sum_{i=1}^{\mathrm{E}}\pi_i\prod_{j=1}^{c}\mu_{i,j}
\end{equation}    
\end{lemma}

The proof of this lemma can refer to Theorem 9 of \cite{allman2009identifiability}. In general, Lemma \ref{app:lemma} shows that if the joint distribution of observation $\mathbb{P}$ can be decomposed into three linearly independent measures w.r.t. $\mu_{i,j}$ as shown in Equation (\ref{equ:app_lemma}), then the distributions of discrete latent variables are identifiable. Based on this Lemma, we further show the identification results of latent environments as follows.

\begin{theorem}
\label{the:blockwise}
(\textbf{Block-wise identifiability of the nonstationary latent variables $\rvz_t^e$ and the stationary latent variables $\rvz_t^s$.} ) We follow the data generation process in Figure 2 and Equation (1)-(3), then we make the following assumptions:
\begin{itemize}
    \item A1 (\textbf{Smooth and Positive Density:}) The probability density function of latent variables is smooth and positive, i.e., $p(\rvz^e_t|\rvz_{t-1}^e,\rvz_{t-1}^e)>0$ over $\mathcal{Z}_t^e,\mathcal{Z}_{t-1}^e$ and $\mathcal{Z}_{t-2}^e$.
    \item A2 (\textbf{Linear Independent:}) For any $\rvz_t^e\in\mathcal{Z}_t^e\subseteq \mathbb{R}^{n_e}$, $\rvv_{t-1,1},\cdots,\rvv_{t-1,n_e}$ as $n_e$ vector functions in $z_{t-2,1},\cdots,\rvv_{t-2,l},\cdots,z_{t-2,n_e}$ are linear independent, where $\rvv_{t-2,l}$ are formalized as follows:
    \begin{equation}
        \rvv_{t-2,l}=\frac{\partial \log p(\rvz_t^e|\rvz_{t-1}^e,\rvz_{t-2}^e)}{\partial z_{t,k}^e\partial z_{t-2,l}^s}
    \end{equation}
    \item A3 \textbf{(Domain Variability:}) There exist two values of $\rvu=\{\rvz_{t-1}^e,\rvz_{t-2}^e\}$, i.e., $\rvu_1$ and $\rvu_2$, s.t., for any set $\mathcal{A}_{\rvz_t}\subseteq \mathcal{Z}_t$ with non-zero probability measure and cannot be expressed as $B_{\rvz_t^s}\times \rvz_t^e$, for any $B_{\rvz_t^s} \subset \mathcal{Z}_t^s$, we have:
    \begin{equation}
        \int_{\rvz_t\in A_{\rvz_t}}p(\rvz_t|\rvu_1)d \rvz_t\neq \int_{\rvz_t\in A_{\rvz_t}}p(\rvz_t|\rvu_2)d \rvz_t
    \end{equation}
\end{itemize}
Then, by learning the data generation process, $\rvz_t^e$ are subspace identifiable.
\end{theorem}
\begin{proof}
We start from the matched marginal distribution to develop the relation between $\rvz_t$ and $\hat{\rvz}_t$ as follows
\begin{equation}
\label{equ:the3_1}
\begin{split}
p(\hat{\rvx}_t)=p(\rvx_t) \Longleftrightarrow p(\hat{g}(\hat{\rvz}_t))=p(g(\rvz_t)) &\Longleftrightarrow p(g^{-1}\circ\hat{g}(\hat{\rvz_t}))|\mathbf{J}_{g^{-1}}|=p(\rvz_t)|\mathbf{J}_{g^{-1}}| \Longleftrightarrow \\&p(h(\hat{\rvz}_t))=p(\rvz_t),
\end{split}
\end{equation}
where $\hat{g}^{-1}: \mathcal{X}\rightarrow \mathcal{Z}$ denotes the estimated invertible generation function, and $h:=g^{-1}\circ \hat{g}$ is the transformation between the true latent variables and the estimated one. $|\mathbf{J}_{g^{-1}}|$ denotes the absolute value of Jacobian matrix determinant of $g^{-1}$. Note that as both $\hat{g}^{-1}$ and $g$ are invertible, $|\mathbf{J}_{g^{-1}}|\neq 0$ and $h$ is invertible.

For any $\rvz_{t-1}$ and $\rvz_{t-2}$, the Jacobian matrix of the mapping from $(\rvx_{t-1}, \hat{\rvz}_t)$ to $(\rvx_{t-1}, \rvz_{t})$ is
\[
\begin{bmatrix}
  \mathbf{I} & \mathbf{0} \\
  * & \mathbf{J}_h \\
\end{bmatrix},
\]
where $*$ denotes a matrix, and the determinant of this Jacobian matrix is $|\mathbf{J}_h|$. Since $\rvx_{t-1}$ do not contain any information of $\hat{\rvz}_{t}$, the right-top element is $\mathbf{0}$. Therefore, $p(\hat{\rvz}_t, \rvx_{t-1}|\rvx_{t-2})=p(\rvz_t, \rvx_{t-1}|\rvx_{t-2})\cdot|\mathbf{J}_h|$. Dividing both sides of this equation by $p(\rvx_{t-1}|\rve_t)$ gives 
\begin{equation}
    p(\hat{\rvz}_t|\rvx_{t-1},\rvx_{t-2})=p(\rvz_t|\rvx_{t-1},\rvx_{t-2})\cdot|\mathbf{J}_h|.
\end{equation}
Since $ p({\rvz}_t|\rvx_{t-1},\rvx_{t-2})= p({\rvz}_t|g(\rvz_{t-1}),g(\rvz_{t-2}))=p({\rvz}_t|\rvz_{t-1},\rvz_{t-2})$ and similarly $p(\hat{\rvz}_t|\rvx_{t-1},\rvx_{t-2})=p(\hat{\rvz}_t|\rvz_{t-1},\rvx_{t-2})$, we have:
\begin{equation}
\label{equ:the3_4}
\begin{split}
    \log p(\hat{\rvz}_t|\hat{\rvz}_{t-1}, \rvz_{t-2})&=\log p(\rvz_t|\rvz_{t-1}, \rvz_{t-2}) + \log |\mathbf{J}_h|=\log p(\rvz_t^e|\rvz_{t-1}^e,\rvz_{t-2}^e) + \log p(\rvz_t^s|\rvz_{t-1}^s) + \log |\mathbf{J}_h|.
\end{split}
\end{equation}
Therefore, for $i\in \{n_e+1, \cdots, n\}$, the partial derivative of Equation (\ref{equ:the3_4}) w.r.t $\hat{z}_{t,i}$ is 
\begin{equation}
\begin{split}
    \frac{\partial \log p(\hat{\rvz}_{t}|{\rvz}_{t-1},{\rvz}_{t-2})}{\partial \hat{z}_{t,i}}&=\frac{\partial \log p(\hat{\rvz}_{t}^e|{\rvz}_{t-1}^e,{\rvz}_{t-2}^e)}{\partial \hat{z}_{t,i}}+ \frac{\partial \log p(\hat{\rvz}_{t}^s|{\rvz}_{t-1}^s)}{\partial \hat{z}_{t,i}}\\&=\sum_{k=1}^{n_e}\frac{\partial \log p(\rvz_t^e|\rvz_{t-1}^e,\rvz_{t-2}^e)}{\partial z_{t,k}^e}\cdot\frac{\partial z_{t,k}^e}{\partial \hat{z}_{t,i}} + \sum_{k=n_e+1}^n \frac{\partial \log p(z^s_{t,k}|\rvz_{t-1}^s)}{\partial z_{t,k}^s}\cdot\frac{\partial z_{t,k}^s}{\partial \hat{z}_{t,i}} + \frac{\partial \log |\mathbf{J_h}|}{\partial \hat{z}_{t,i}}.
\end{split}
\end{equation}
    Sequentially, for each $l=1,\cdots,n_e$, and each value of $z_{t-2,l}^e$, its partial derivative w.r.t. $z_{t-2,l}^e$ is shown as follows:
\begin{equation}
\label{equ:the3_5}
\begin{split}
    &\frac{\partial \log p(\hat{\rvz}_{t}|{\rvz}_{t-1},{\rvz}_{t-2})}{\partial \hat{z}_{t,i}\partial z_{t-2,l}^s}=\frac{\partial \log p(\hat{\rvz}_{t}^e|{\rvz}_{t-1}^e,{\rvz}_{t-2}^e)}{\partial \hat{z}_{t,i}\partial z_{t-2,l}^s}+ \frac{\partial \log p(\hat{\rvz}_{t}^s|{\rvz}_{t-1}^s)}{\partial \hat{z}_{t,i}\partial z_{t-2,l}^s}\\=&\sum_{k=1}^{n_e}\frac{\partial \log p(\rvz_t^e|\rvz_{t-1}^e,\rvz_{t-2}^e)}{\partial z_{t,k}^e\partial z_{t-2,l}^s}\cdot\frac{\partial z_{t,k}^e}{\partial \hat{z}_{t,i}} + \sum_{k=n_e+1}^n \frac{\partial \log p(z^s_{t,k}|\rvz_{t-1}^s)}{\partial z_{t,k}^s\partial z_{t-2,l}^s}\cdot\frac{\partial z_{t,k}^s}{\partial \hat{z}_{t,i}} + \frac{\partial \log |\mathbf{J_h}|}{\partial \hat{z}_{t,i}\partial z_{t-2,l}^s}.
\end{split}
\end{equation}
Since the distribution $p(\hat{\rvz}_t^e|\rvz_{t-1}^e, \rvz_{t-2}^e)$ does not change across $\hat{z}_{t,i}, i\in \{n_e+1,\cdots,n\}$, $\frac{\partial \log p(\hat{\rvz}_t^e|\rvz_{t-1}^e, \rvz_{t-2}^e)}{\partial \hat{z}_{t,i}}=0$. Since the distribution $p(\hat{\rvz}_t^s|\rvz_{t-1}^s)$ does not change across different value of $z_{t-2,l}^e$, 
$\frac{\partial \log p(\hat{\rvz}_{t}^s|{\rvz}_{t-1}^s)}{\partial \hat{z}_{t,i}\partial z_{t-2,l}}=0$. And given $\rvz_{t-1}^s$, $\rvz_{t-2}$ is independent of $\rvz_t^s$, so $\frac{\partial\log p(z_{t,k}^s|\rvz_{t-1}^s)}{\partial z_{t,k}^s\partial z_{t-2,l}^s}=0$. Moreover, $\frac{\partial \log |\mathbf{J_h}|}{\partial \hat{z}_{t,i}\partial z_{t-2,l}^s}=0$, then Equation (\ref{equ:the3_5}) can be rewritten as:
\begin{equation}
    0=\sum_{k=1}^{n_e}\frac{\partial \log p(\rvz_t^e|\rvz_{t-1}^e,\rvz_{t-2}^e)}{\partial z_{t,k}^e\partial z_{t-2,l}^s}\cdot\frac{\partial z_{t,k}^e}{\partial \hat{z}_{t,i}}
\end{equation}

Based on the linear independence assumption A1, the linear system is a $n_e\times n_e$ full-rank system. Therefore, the only solution is $\frac{\partial z_{t,k}^e}{\partial \hat{z}_{t,i}}=0$ for $i=\{n_e+1,\cdots,n\}$ and $k\in\{1, \cdots, n_e\}$. Since $h(\cdot)$ is smooth over $\mathcal{Z}$, its Jacobian can be formalized as follows:
\begin{equation}
\begin{gathered}\nonumber
    \mathbf{J}_h=\begin{bmatrix}
    \begin{array}{c|c}
        \textbf{A}:=\frac{\partial \rvz^e_t}{\partial \hat{\rvz}^e_t} & \textbf{B}:=\frac{\partial \rvz_t^e}{\partial \hat{\rvz}_t^s} \\ \midrule 
        \textbf{C}:=\frac{\partial \rvz^s_t}{\partial \hat{\rvz}^e_t} & \textbf{D}:=\frac{\partial \rvz_t^s}{\partial \hat{\rvz}_t^s}
    \end{array}
    \end{bmatrix}.
\end{gathered}
\end{equation}
Note that $\frac{\partial z_{t,k}^e}{\partial \hat{z}_{t,i}}=0$ for $i=\{n_e+1,\cdots,n\}$ and $k\in\{1, \cdots, n_d\}$ means $\mathbf{B}=0$. Since $h(\cdot)$ is invertible, $\mathbf{J}_h$ is a full-rank matrix. Therefore, $\textbf{A}\neq 0$. 

Besides, based on A3, one can show that all entries in the submatrix C zero according to part of the proof of Theorem 4.2 in \cite{kong2022partial}(Steps 1, 2, and 3). Therefore, $\rvz_t^s$ and \(\rvz_t^{e}\) are block-wise identifiable.
\end{proof}

\begin{theorem}
\textbf{(Identifiability of the latent environment $e_t.$)} Suppose the observed data is generated following the data
generation process in Figure 3 and Equation (1)-(3). Then we further make the following assumptions:
\begin{itemize}
    \item A4 (\underline{\textit{Prior Environment Number:}}) The number of latent environments of the Markov process, $E$, is known.
    \item A5 (\underline{\textit{Full Rank:}}) The transition matrix $\mathbf{A}$ is full rank.
    \item A6 (\underline{\textit{Linear Independence:}}) For $e=1,2,\cdots, E$, the probability measures $\mu_e=p(\rvz_t^e|e_t)$ are linearly independence and for any two different probability measures $\mu_i, \mu_j$, their ratio $\frac{\mu_{i}}{\mu_{j}}$ are linearly independence.
\end{itemize}
Then, by modeling the observations $\rvx_1,\rvx_2,\cdots, \rvx_t$, the joint distribution of the corresponding latent environment variables $p(\rve_1, \rve_2, \cdots, \rve_t)$ is identifiable up to label swapping of the hidden environment.
\end{theorem}
\begin{proof}
Suppose we have:
\begin{equation}
    \hat{p}(\rvx_1, \rvx_2, \cdots, \rvx_T)=p(\rvx_1, \rvx_2, \cdots, \rvx_T),
\end{equation}
where $\hat{p}(\rvx_1,\rvx_2,\cdots,\rvx_T)$ and $p(\rvx_1,\rvx_2,\cdots,\rvx_T)$ denote the estimated and ground-truth joint distributions, respectively; and $p(\rvx_1, \rvx_2, \cdots, \rvx_T)$ has transition matrix $\mathbf{A}$ and emission distribution $(\mu_1, \cdots, \mu_E)$, similarly for $\hat{p}(\rvx_1, \rvx_2, \cdots, \rvx_T)$.

According to Theorem 1, since the nonstationary latent variables are block-wise identifiable, we can consider three consecutive nonstationary latent variables $\rvz_1^e,\rvz_2^e,\rvz_3^e$ and corresponding three discrete elements $e_1,e_2,e_3$.

\begin{equation}
\begin{split}
&p(\rvz_1^e,\rvz_2^e,\rvz_3^e)=\sum_{e_1,e_2,e_3}p(\rvz_1^e,\rvz_2^e,\rvz_3^e,e_1,e_2,e_3)=\sum_{e_1,e_2,e_3}p(e_2)p(\rvz_1^e,\rvz_2^e,\rvz_3^e,e_1,e_3|e_2)\\=&\sum_{e_1,e_2,e_3}p(e_2)p(z_2^e|e_2)p(\rvz_1^e,\rvz_3^e,e_1,e_3|e_2,\rvz_2^e)=\sum_{e_1,e_2,e_3}p(e_2)p(z_2^e|e_2)p(\rvz_1^e,e_1|e_2)p(\rvz_3^e, e_3|e_2)\\=& \sum_{e_1,e_2,e_3}p(e_2)p(z_2^e|e_2)p(\rvz_1^e|e_1)p(e_1|e_2)p(\rvz_3^e|e_3)p(e_3|e_2)\\=& \sum_{e_2}p(e_2)\underbrace{\Big(\sum_{e_1}p(\rvz_1^e|e_1)p(e_1|e_2)\Big)}_{\bar{\mu}_{e_2}}\cdot \mu_{e_2} \cdot \underbrace{\Big(\sum_{e_3}p(\rvz_3^e|e_3)p(e_3|e_2)\Big)}_{\dot{\mu}_{e_2}}.
\end{split}
\end{equation}
According A5 and A6, $\mathbf{A}$ is full rank and the probability measure $\mu_1,\mu_2,\cdots,\mu_E$ are linearly independent, the probability measure $\overline{\mu}_{e_2}=\sum_{\rve_1}A_{e_2,e_1}\cdot\mu_{e_2}$ are linearly independent and the probability measure $\dot{\mu}_{e_2}=\sum_{e_3}A_{e_2,e_3}\cdot \mu_{e_2}$ are also linearly independent,
Thus, applying Theorem 9 of \cite{allman2009identifiability}, there exists a permutation $\sigma$ of $\{1,\cdots, E\}$, such that, $\forall i\in \{1,\cdots,E\}$:
\begin{equation}
\begin{split}
    \tilde{\mu}_i&=\mu_{\sigma(i)}\\
\sum_j\tilde{A}_{i,j}\tilde{\mu}_i&=\sum_jA_{\sigma(i),j}\mu_{i}
\end{split}
\end{equation}
This gives easily $\forall i\in \{1,\cdots,E\}$, we can obtain:
\begin{equation}
\begin{split}
    \sum_j \tilde{A}_{i,j}\mu_{\sigma(j)}=\sum_j A_{\sigma(i), \sigma(j)}\mu_{\sigma(j)}.
\end{split}
\end{equation}
Since the $\mu_j$ is linearly independent, we can establish the equivalence between $\tilde{\mathbf{A}}$ and $\mathbf{A}$ via permutation $\sigma$, i.e., $\tilde{A}{i,j}=A_{\sigma(j),\sigma(j)}$,
\end{proof}

\subsection{Component-wise Identification of Stationary Latent Variables $\rvz_t^s$}\label{app:the2}
    
\begin{proof}
We start from the matched marginal distribution to develop the relation between $\rvz$ and $\hat{\rvz}$ as follows
\begin{equation}
\label{equ:the2_1}
\begin{split}
p(\hat{\rvx}_t)=p(\rvx_t) \Longleftrightarrow p(\hat{g}(\hat{\rvz}_t))=p(g(\rvz_t)) &\Longleftrightarrow p(g^{-1}\circ\hat{g}(\hat{\rvz_t}))|\mathbf{J}_{g^{-1}}|=p(\rvz_t)|\mathbf{J}_{g^{-1}}| \Longleftrightarrow \\&p(h(\hat{\rvz}_t))=p(\rvz_t),
\end{split}
\end{equation}
where $\hat{g}^{-1}: \mathcal{X}\rightarrow \mathcal{Z}$ denotes the estimated invertible generation function, and $h:=g^{-1}\circ \hat{g}$ is the transformation between the true latent variables and the estimated one. $|\mathbf{J}_{g^{-1}}|$ denotes the absolute value of Jacobian matrix determinant of $g^{-1}$. Note that as both $\hat{g}^{-1}$ and $g$ are invertible, $|\mathbf{J}_{g^{-1}}|\neq 0$ and $h$ is invertible.

Then for any $\rvu_t$, the Jacobian matrix of the mapping from $(\rvx_{t-1}, \hat{\rvz}_t)$ to $(\rvx_{t-1}, \rvz_{t})$ is
\[
\begin{bmatrix}
  \mathbf{I} & \mathbf{0} \\
  * & \mathbf{J}_h \\
\end{bmatrix},
\]
where $*$ denotes a matrix, and the determinant of this Jacobian matrix is $|\mathbf{J}_h|$. Since $\rvx_{t-1}$ do not contain any information of $\hat{\rvz}_{t}$, the right-top element is $\mathbf{0}$.
Therefore $p(\hat{\rvz}_t, \rvx_{t-1}|\rve_t)=p(\rvz_t, \rvx_{t-1}|\rve_t)\cdot |\mathbf{J}_h|$. Dividing both sides of this equation by $p(\rvx_{t-1}|\rve_t)$ gives

\begin{equation}
    p(\hat{\rvz}_t|\rvx_{t-1}, \rve_t)=p(\rvz_t|\rvx_{t-1},\rve_t)\cdot |\mathbf{J}_h|
\end{equation}
Since $p(\rvz_t|\rvz_{t-1}, \rve_t)=p(\rvz_t|g(\rvz_{t-1}), \rve_t)=p(\rvz_t|\rvx_{t-1},\rve_t)$ and similarly $p(\hat{\rvz}_{t}|\hat{\rvz}_{t-1}, \rve_t)=p(\hat{\rvz}_{t}|\rvx_{t-1}, \rve_t)$, we have
\begin{equation}
\label{equ:the2_4}
\begin{split}
    \log p(\hat{\rvz}_t|\hat{\rvz}_{t-1}, \rve_t)&=\log p(\rvz_t|\rvz_{t-1}, \rve_t) + \log |\mathbf{J}_h|=\sum_{k=1}^n \log p(z_{t,k}|\rvz_{t-1},\rve_t) + \log |\mathbf{J}_h|\\&=\sum_{k=1}^{n_e} \log p(z_{t,k}^e|\rve_t) + \sum_{k=n_e+1}^n \log p(z_{t,k}^s|\rvz_{t-1}^s) + \log |\mathbf{J}_h|.
\end{split}
\end{equation}
Therefore, for $i\in \{n_e+1, \cdots, n\}$, the partial derivative of Equation (23) w.r.t. $\hat{z}_{t,i}$ is 
\begin{equation}
\label{equ:the2_5}
    \frac{\partial \log p(\hat{z}_{t,i}|\hat{\rvz}_{t-1},\rve_t)}{\partial \hat{z}_{t,i}}=\sum_{k=1}^{n_e}\frac{\partial \log p(z_{t,k}^e|\rve_t)}{\partial z_{t,k}^e}\cdot \frac{\partial z_{t,k}^e}{\partial \hat{z}_{t,i}} + \sum_{k=n_e+1}^n\frac{\partial \log p(z_{t,k}^s|\rvz_{t-1}^s)}{\partial z_{t,k}^s}\cdot \frac{\partial z_{t,k}^s}{\partial \hat{z}_{t,i}} + \frac{\partial \log |\mathbf{J}_h|}{\partial \hat{z}_{t,i}},
\end{equation}

And for $j\in \{n_e+1,\cdots,n\}$, the second-order derivative of Equation (\ref{equ:the2_5}) is 
\begin{equation}
\small
\begin{split}
0=\frac{\partial \log p(\hat{z}_{t,i}|\hat{\rvz_{t-1,e_t}})}{\partial \hat{z}_{t,i}\partial \hat{z}_{t,j}}&=\sum_{k=1}^{n_e}\Big(\frac{\partial \log p(z_{t,k}^e|e_t)}{\partial^2 z_{t,k}^e}\cdot\frac{\partial z_{t,k}^e}{\partial \hat{z}_{t,i}}\cdot\frac{\partial z_{t,k}^e}{\partial \hat{z}_{t,j}}+\frac{\partial \log p(z_{t,k}^e|e_t)}{\partial z_{t,k}^e}\cdot\frac{\partial^2 z_{z,k}^e}{\partial \hat{z}_{t,i}\partial \hat{z}_{t,j}}\Big) + \\\sum_{k=n_e+1}^n\Big(&\frac{\partial \log p(z_{t,k}^s|\rvz_{t-1}^s)}{\partial^2 z_{t,k}^s}\cdot\frac{\partial z_{t,k}^s}{\partial \hat{z}_{t,i}}\cdot\frac{\partial z_{t,k}^s}{\partial \hat{z}_{t,j}}+\frac{\partial \log p(z_{t,k}^s|\rvz_{t-1}^s)}{\partial z_{t,k}^s}\cdot\frac{\partial^2 z_{t,k}^s}{\partial \hat{z}_{t,i}\partial \hat{z}_{t,j}}\Big)+\frac{\partial^2 \log |\mathbf{J}_h|}{\partial \hat{z}_{t,i}\partial \hat{z}_{t,j}}
\end{split}
\end{equation}

For each $l =n_e+1,\cdots n$ and each value of $z_{t-1,l}$, its partial derivative w.r.t. $z_{t-1,l}$ is shown as follows
\begin{equation}
\small
\begin{split}
0=\frac{\partial^3 \log p(\hat{z}_{t,i}|\hat{\rvz_{t-1,e_t}})}{\partial \hat{z}_{t,i}\partial \hat{z}_{t,j} \partial z_{t-1,l}}&=\sum_{k=1}^{n_e}\Big(\frac{\partial^3 \log p(z_{t,k}^e|e_t)}{\partial^2 z_{t,k}^e \partial z_{t-1,l}}\cdot\frac{\partial z_{t,k}^e}{\partial \hat{z}_{t,i}}\cdot\frac{\partial z_{t,k}^e}{\partial \hat{z}_{t,j}}+\frac{\partial^2 \log p(z_{t,k}^e|e_t)}{\partial z_{t,k}^e\partial z_{t-1,l}}\cdot\frac{\partial^2 z_{z,k}^e}{\partial \hat{z}_{t,i}\partial \hat{z}_{t,j}}\Big) + \\\sum_{k=n_e+1}^n\Big(&\frac{\partial^3 \log p(z_{t,k}^s|\rvz_{t-1}^s)}{\partial^2 z_{t,k}^s\partial z_{t-1,l}}\cdot\frac{\partial z_{t,k}^s}{\partial \hat{z}_{t,i}}\cdot\frac{\partial z_{t,k}^s}{\partial \hat{z}_{t,j}}+\frac{\partial^2 \log p(z_{t,k}^s|\rvz_{t-1}^s)}{\partial z_{t,k}^s\partial z_{t-1,l}}\cdot\frac{\partial^2 z_{t,k}^s}{\partial \hat{z}_{t,i}\partial \hat{z}_{t,j}}\Big)+\frac{\partial^3 \log |\mathbf{J}_h|}{\partial \hat{z}_{t,i}\partial \hat{z}_{t,j}\partial z_{t-1,l}}
\end{split}
\end{equation}

Since the distribution $p(z_{t,k}^e|e_t)$ is not influenced by $z_{t-1,l}$, $\frac{\partial^3 \log p(z_{t,k}^e|e_t)}{\partial^2 z_{t,k}^e \partial z_{t-1,l}}=0$ and $\frac{\partial^2 \log p(z_{t,k}^e|e_t)}{\partial z_{t,k}^e\partial z_{t-1,l}}=0$. Moreover, since $\log|\mathbf{J}_h|$ does not depend on $z_{t-1,l}$, $\frac{\partial^3 \log |\mathbf{J}_h|}{\partial \hat{z}_{t,i}\partial \hat{z}_{t,j}\partial z_{t-1,l}}=0$, and the aforementioned equation can be further rewritten as:
\begin{equation}
\small
0=\sum_{k=n_e+1}^n\Big(\frac{\partial^3 \log p(z_{t,k}^s|\rvz_{t-1}^s)}{\partial^2 z_{t,k}^s\partial z_{t-1,l}}\cdot\frac{\partial z_{t,k}^s}{\partial \hat{z}_{t,i}}\cdot\frac{\partial z_{t,k}^s}{\partial \hat{z}_{t,j}}+\frac{\partial^2 \log p(z_{t,k}^s|\rvz_{t-1}^s)}{\partial z_{t,k}^s\partial z_{t-1,l}}\cdot\frac{\partial^2 z_{t,k}^s}{\partial \hat{z}_{t,i}\partial \hat{z}_{t,j}}\Big)
\end{equation}


\begin{equation}
    \frac{\partial \log p(\hat{z}_{t,i}|\hat{\rvz}_{t-1},\rve_t)}{\partial \hat{z}_{t,i}\partial z_{t-1,l}^c}=\sum_{k=n_e+1}^n\left(\frac{\partial \log p(z_{t,k}^s|\rvz_{t-1}^s)}{\partial z_{t,k}^s\partial z_{t-1,l}^s}\cdot \frac{\partial z_{t,k}^s}{\partial \hat{z}_{t,i}} \right) + \frac{\partial \log |\mathbf{J}_h|}{\partial \hat{z}_{t,i}\partial z_{t-1,l}^s},
\end{equation}
Then we subtract the Equation (25) corresponding to $z_{t-1,l}^s$ with that corresponding to $z_{t-1, n}$, and we have:
\begin{equation}
\begin{split}
    \frac{\partial \log p(\hat{z}_{t,i}|\hat{\rvz}_{t-1},\rve_t)}{\partial \hat{z}_{t,i}\partial z_{t-1,l}^s}&-\frac{\partial \log p(\hat{z}_{t,i}|\hat{\rvz}_{t-1},\rve_t)}{\partial \hat{z}_{t,i}\partial z_{t-1,n}^s}\\&=\sum_{k=n_e+1}^n\left((\frac{\partial \log p(z_{t,k}^s|\rvz_{t-1}^s)}{\partial z_{t,k}^s\partial z_{t-1,l}^s} - \frac{\partial \log p(z_{t,k}^s|\rvz_{t-1}^s)}{\partial z_{t,k}^s\partial z_{t-1,n}^s})\cdot \frac{\partial z_{t,k}^s}{\partial \hat{z}_{t,i}} \right) + \frac{\partial \log |\mathbf{J}_h|}{\partial \hat{z}_{t,i}\partial z_{t-1,l}^s}-\frac{\partial \log |\mathbf{J}_h|}{\partial \hat{z}_{t,i}\partial z_{t-1,n}^s}
\end{split}
\end{equation}
Since the distribution of $\hat{z}_{t,i}$ does not change with the $z_{t-1,l}^s$, $\frac{\partial \log p(\hat{z}_{t,i}|\hat{\rvz}_{t-1},\rve_t)}{\partial \hat{z}_{t,i}\partial z_{t-1,l}^s}-\frac{\partial \log p(\hat{z}_{t,i}|\hat{\rvz}_{t-1},\rve_t)}{\partial \hat{z}_{t,i}\partial z_{t-1,n}^s}=0$. Moreover, $\frac{\partial \log |\mathbf{J}_h|}{\partial \hat{z}_{t,i}\partial z_{t-1,l}^s}=0$ for any $l$. Therefore, Equation (26) can be written as follows:
\begin{equation}
    0=\sum_{k=n_e+1}^n \left(\frac{\partial \log p(z_{t,k}^s|\rvz_{t-1}^s)}{\partial z_{t,k}^s\partial z_{t-1,l}^s} - \frac{\partial \log p(z_{t,k}^s|\rvz_{t-1}^s)}{\partial z_{t,k}^s\partial z_{t-1,n}^s}\right)\cdot \frac{\partial z_{t,k}^s}{\partial \hat{z}_{t,i}} 
\end{equation}
According to the linear independence assumption, there is only one solution $\frac{\partial z_{t,k}^s}{\partial \hat{z}_{t,i}}$, meaning that $\textbf{C}$ in the following Jacobian Matrix is 0.
\begin{equation}
\begin{gathered}\nonumber
    \mathbf{J}_h=\begin{bmatrix}
    \begin{array}{c|c}
        \textbf{A}:=\frac{\partial \rvz^e_t}{\partial \hat{\rvz}^e_t} & \textbf{B}:=\frac{\partial \rvz_t^e}{\partial \hat{\rvz}_t^s} \\ \midrule 
        \textbf{C}:=\frac{\partial \rvz^s_t}{\partial \hat{\rvz}^e_t} & \textbf{D}:=\frac{\partial \rvz_t^s}{\partial \hat{\rvz}_t^s}
    \end{array}
    \end{bmatrix}.
\end{gathered}
\end{equation}
Since $h$ is invertible and $\mathbf{J}_h$ is full-rank, for each $z^s_{t,k}$, there exists a $h_k$ such that $z^s_{t,k}=h_k(\hat{\rvz}_{t,i}), i\in n_e+1,\cdots,n$, implying that $\rvz^s_t$ is subspace identifiable.
\end{proof}


\subsection{Component-wise Identification of Nonstationary Latent Variables $\rvz_{t}^e$}\label{app:the3}

\label{the:the2}

We start from the matched marginal distribution to develop the relation between $\rvz$ and $\hat{\rvz}$ as follows
\begin{equation}
\label{equ:the2_1}
\begin{split}
p(\hat{\rvx}_t)=p(\rvx_t) \Longleftrightarrow p(\hat{g}(\hat{\rvz}_t))=p(g(\rvz_t)) &\Longleftrightarrow p(g^{-1}\circ\hat{g}(\hat{\rvz_t}))|\mathbf{J}_{g^{-1}}|=p(\rvz_t)|\mathbf{J}_{g^{-1}}| \Longleftrightarrow \\&p(h(\hat{\rvz}_t))=p(\rvz_t),
\end{split}
\end{equation}
where $\hat{g}^{-1}: \mathcal{X}\rightarrow \mathcal{Z}$ denotes the estimated invertible generation function, and $h:=g^{-1}\circ \hat{g}$ is the transformation between the true latent variables and the estimated one. $|\mathbf{J}_{g^{-1}}|$ denotes the absolute value of Jacobian matrix determinant of $g^{-1}$. Note that as both $\hat{g}^{-1}$ and $g$ are invertible, $|\mathbf{J}_{g^{-1}}|\neq 0$ and $h$ is invertible.

First, it is straightforward to find that if the components of $\hat{\rvz}_t$ are mutually independent conditional on previous $\hat{\rvz}_t$ and current $\rve_t$, then for any $i\neq j$, $\hat{z}_{t,i}$ and $\hat{z}_{t,j}$ are conditionally independent given $\hat{\rvz}_{t-1} \cup (\hat{\rvz}\setminus \{\hat{z}_{t,i}, \hat{z}_{t,j}\}, \rve_t)$, i.e.
\begin{equation}
    p(\hat{z}_{t,i}|\hat{\rvz}_{t-1}, \rve_t)=p(\hat{z}_{t,i}|\hat{\rvz}_{t-1} \setminus \{\hat{z}_{t,i}, \hat{z}_{t,j}\}, \rve_t).
\end{equation}
At the same time, it also implies $\hat{z}_{t,i}$ is independent from $\hat{\rvz}_{t}\setminus \{\hat{z}_{t,i}\}$ conditional on $\hat{\rvz}_{t-1}$ and $\rve_t$, i.e.,
\begin{equation}
    p(\hat{z}_{t,i}|\hat{\rvz}_{t-1}, \rve_t)=p(\hat{z}_{t,i}|\hat{\rvz}_{t-1} \setminus \{\hat{z}_{t,i}\}, \rve_t).
\end{equation}

Combining the above two equations gives
\begin{equation}
    p(\hat{z}_{t,i}|\hat{\rvz}_{t-1}\cup (\hat{\rvz}_t \setminus \{\hat{z}_{t,i}\}), \rve_t)=p(\hat{z}_{t,i}|\hat{\rvz}_{t-1}\cup (\hat{\rvz}_t \setminus \{\hat{z}_{t,i}, \hat{z}_{t,j}\}), \rve_t),
\end{equation}
i.e., for $i\neq j$, $\hat{z}_{t,i}$ and $\hat{z}_{t,j}$ are conditionally independent given $\hat{\rvz}_{t-1}\cup (\hat{\rvz}_t \setminus \{\hat{z}_{t,i}, \hat{z}_{t,j}\})\cup\{\rve_t\}$, which implies that
\begin{equation}
    \frac{\partial^2 \log p(\hat{\rvz}_t, \hat{\rvz}_{t-1},\rve_t)}{\partial \hat{z}_{t,i} \partial \hat{z}_{t,j}}=0,
\end{equation}
Assuming that the cross second-order derivative exists \cite{spantini2018inference}. Since $p(\hat{\rvz}_t, \hat{\rvz}_{t-1},\rve_t)=p(\hat{\rvz}_t|\hat{\rvz}_{t-1},\rve_t)p(\hat{\rvz}_{t-1}, \rve_t)$ while $p(\hat{\rvz}_{t-1}, \rve_t)$ does not involve $\hat{z}_{t,i}$ and $\hat{z}_{t,j}$, the above equality is equivalent to
\begin{equation}
    \frac{\partial^2 \log p(\hat{\rvz}_t| \hat{\rvz}_{t-1},\rve_t)}{\partial \hat{z}_{t,i} \partial \hat{z}_{t,j}}=0.
\end{equation}
Then for any $\rve_t$, the Jacobian matrix of the mapping from $(\rvx_{t-1}, \hat{\rvz}_t)$ to $(\rvx_{t-1}, \rvz_{t})$ is
\[
\begin{bmatrix}
  \mathbf{I} & \mathbf{0} \\
  * & \mathbf{J}_h \\
\end{bmatrix},
\]
where $*$ denotes a matrix, and the determinant of this Jacobian matrix is $|\mathbf{J}_h|$. Since $\rvx_{t-1}$ do not contain any information of $\hat{\rvz}_{t}$, the right-top element is $\mathbf{0}$.
Therefore $p(\hat{\rvz}_t, \rvx_{t-1}|\rve_t)=p(\rvz_t, \rvx_{t-1}|\rve_t)\cdot |\mathbf{J}_h|$. Dividing both sides of this equation by $p(\rvx_{t-1}|\rve_t)$ gives

\begin{equation}
    p(\hat{\rvz}_t|\rvx_{t-1}, \rve_t)=p(\rvz_t|\rvx_{t-1},\rve_t)\cdot |\mathbf{J}_h|
\end{equation}

Since $p(\rvz_t|\rvz_{t-1}, \rve_t)=p(\rvz_t|g(\rvz_{t-1}), \rve_t)=p(\rvz_t|\rvx_{t-1},\rve_t)$ and similarly $p(\hat{\rvz}_{t}|\hat{\rvz}_{t-1}, \rve_t)=p(\hat{\rvz}_{t}|\rvx_{t-1}, \rve_t)$, we have
\begin{equation}
\label{equ:the2_4}
\begin{split}
    \log p(\hat{\rvz}_t|\hat{\rvz}_{t-1}, \rve_t)&=\log p(\rvz_t|\rvz_{t-1}, \rve_t) + \log |\mathbf{J}_h|=\sum_{k=1}^n \log p(z_{t,k}|\rvz_{t-1},\rve_t) + \log |\mathbf{J}_h|\\&=\sum_{k=1}^{n_e} \log p(z_{t,k}^e|\rve_t) + \sum_{k=n_e+1}^n \log p(z_{t,k}^s|\rvz_{t-1}^s) + \log |\mathbf{J}_h|.
\end{split}
\end{equation}
Therefore, for $i\in \{n_e+1, \cdots, n\}$, the partial derivative of Equation (\ref{equ:the2_4}) w.r.t. $\hat{z}_{t,i}$ is 
\begin{equation}
\label{equ:the2_5}
    \frac{\partial \log p(\hat{z}_{t,i}|\hat{\rvz}_{t-1},\rve_t)}{\partial \hat{z}_{t,i}}=\sum_{k=1}^{n_e}\frac{\partial \log p(z_{t,k}^e|\rve_t)}{\partial z_{t,k}^e}\cdot \frac{\partial z_{t,k}^e}{\partial \hat{z}_{t,i}} + \sum_{k=n_e+1}^n\frac{\partial p(z_{t,k}^s|\rvz_{t-1}^s)}{\partial z_{t,k}^s}\cdot \frac{\partial z_{t,k}^s}{\partial \hat{z}_{t,i}} + \frac{\partial \log |\mathbf{J}_h|}{\partial \hat{z}_{t,i}},
\end{equation}
Suppose $\rvu_t=e_1, \cdots, e_{|E|}$, we subtract the Equation (\ref{equ:the2_5}) corresponding to $e_l$ with that corresponds to $e_0$ and have:
\begin{equation}
\label{equ:the2_6}
    \sum_{k=1}^{n_e}\left( \frac{\partial \log p(z_{t,k}^e|\rve_l)}{\partial z_{t,k}^e} -\frac{\partial \log p(z_{t,k}^e|\rve_0)}{\partial z_{t,k}^e}\right) \cdot \frac{\partial z_{t,k}^e}{\partial \hat{z}_{t,i}}=\frac{\partial \log p(\hat{z}_{t,i}|\hat{\rvz}_{t-1},\rve_l)}{\partial \hat{z}_{t,i}}-\frac{\partial \log p(\hat{z}_{t,i}|\hat{\rvz}_{t-1},\rve_0 )}{\partial \hat{z}_{t,i}}.
\end{equation}

Since the distribution of estimated $\hat{z}_{t,i}$ does not change across different domains, $\frac{\partial \log p(\hat{z}_{t,i}|\hat{\rvz}_{t-1},\rve_l)}{\partial \hat{z}_{t,i}}-\frac{\partial \log p(\hat{z}_{t,i}|\hat{\rvz}_{t-1},\rve_0 )}{\partial \hat{z}_{t,i}}=0$. Since $\sum_{k=n_e+1}^n\frac{\partial p(z_{t,k}^s|\rvz_{t-1}^s)}{\partial z_{t,k}^s}\cdot \frac{\partial z_{t,k}^s}{\partial \hat{z}_{t,i}} + \frac{\partial \log |\mathbf{J}_h|}{\partial \hat{z}_{t,i}}$ does not change across domains, we have 
\begin{equation}
\label{equ:the2_7}
    \sum_{k=1}^{n_e}\left( \frac{\partial \log p(z_{t,k}^e|\rve_l)}{\partial z_{t,k}^e} -\frac{\partial \log p(z_{t,k}^e|\rve_0)}{\partial z_{t,k}^e}\right) \cdot \frac{\partial z_{t,k}^e}{\partial \hat{z}_{t,i}}=0.
\end{equation}

Based on the linear independence assumption A7, the linear system is a $n_e\times n_e$ full-rank system. Therefore, the only solution is $\frac{\partial z_{t,k}^e}{\partial \hat{z}_{t,i}}=0$ for $i=\{n_e+1,\cdots,n\}$ and $k\in\{1, \cdots, n_e\}$. Since $h(\cdot)$ is smooth over $\mathcal{Z}$, its Jacobian can be formalized as follows:
\begin{equation}
\begin{gathered}\nonumber
    \mathbf{J}_h=\begin{bmatrix}
    \begin{array}{c|c}
        \textbf{A}:=\frac{\partial \rvz^e_t}{\partial \hat{\rvz}^e_t} & \textbf{B}:=\frac{\partial \rvz_t^e}{\partial \hat{\rvz}_t^s} \\ \midrule 
        \textbf{C}:=\frac{\partial \rvz^s_t}{\partial \hat{\rvz}^e_t} & \textbf{D}:=\frac{\partial \rvz_t^s}{\partial \hat{\rvz}_t^s}
    \end{array}
    \end{bmatrix}.
\end{gathered}
\end{equation}
Note that $\frac{\partial z_{t,k}^e}{\partial \hat{z}_{t,i}}=0$ for $i=\{n_e+1,\cdots,n\}$ and $k\in\{1, \cdots, n_d\}$ means $\mathbf{B}=0$. Since $h(\cdot)$ is invertible, $\mathbf{J}_h$ is a full-rank matrix. Therefore, for each $z^e_{t,i}, i\in \{1,\cdots,n_e\}$, there exists a $h_i$ such that $z^e_{t,i}=h_i(\hat{\rvz}^e)$.


\section{More Details of Experiment}\label{app:exp_detail}



\subsection{Simulation Experiments} \label{app:simulation}

To validate if the proposed method can reconstruct the Markov transition matrix and infer the latent environments, we examine the accuracy of estimating latent environments, which is shown in Table \ref{tab:env_acc}. 
We consider the Mean Square Error (MSE) between the ground truth \textbf{A} and the estimated one and the accuracy of estimating $\hat{e}_t$ to evaluate how well the proposed method can estimate latent environments. Note that the MSE and Accuracy are influenced by the permutation, which is similar to the clustering evaluation problems, so we explored all permutations and selected the best
possible assignment for evaluation.
According to the experiment results, we can find that the proposed method can identify the latent environment with high accuracy, which is consistent with the theory. 

\begin{table}[h]
\centering%
\caption{Experiment results of two synthetic datasets on estimating environment indices}
\begin{tabular}{@{}c|c|c@{}}
\toprule
        & \multicolumn{2}{c}{Unknown Nonstationary Metrics} \\ \midrule
Dataset & Accuracy Estimating $\rvu_t$       & MSE estimating       \\ \midrule
A       & 91.9                       & 0.0103               \\
B       & 85.8                       & 0.0163               \\ \bottomrule
\end{tabular}%
\label{tab:env_acc}
\end{table}

\subsection{Real-World Datasets}
\subsubsection{Dataset Description}\label{app:dataset}
\begin{itemize}
    \item \textbf{ETT} \cite{zhou2021informer} is an electricity transformer temperature dataset collected from two separated counties in China, which contains two separate datasets $\{\text{ETTh1, ETTh2}\}$ for one hour level. 
    \item \textbf{Exchange} \cite{lai2018modeling} is the daily exchange rate dataset from of eight foreign countries including Australia, British, Canada, Switzerland, China, Japan, New Zealand, and Singapore ranging from 1990 to.
    \item \textbf{ILI} \footnote{https://gis.cdc.gov/grasp/fluview/fluportaldashboard.html} is a real-world public dataset of influenza-like illness, which records weekly influenza activity levels (measured by the weighted ILI metric) in 10 districts (divided by HHS) of the mainland United States between the first week of 2010 and the 52nd week of 2016.
    \item \textbf{Weather }\footnote{https://www.bgc-jena.mpg.de/wetter/} is recorded at the Weather Station at the Max Planck Institute
for Biogeochemistry in Jena, Germany.
    \item \textbf{ECL} \footnote{https://archive.ics.uci.edu/dataset/321/electricityloaddiagrams20112014} is an electricity consuming load dataset with the electricity consumption (kWh) collected from 321 clients. 
    \item \textbf{Traffic} \footnote{https://pems.dot.ca.gov/} is a dataset of traffic speeds collected from the California Transportation Agencies (CalTrans) Performance Measurement System (PeMS), which contains data collected from 325 sensors located throughout the Bay Area.
    \item \textbf{M4} dataset \cite{makridakis2020m4} is a collection of 100,000 time series used for the fourth edition of the Makridakis forecasting Competition with time series of yearly, quarterly, monthly and other (weekly, daily and hourly) data.
\end{itemize}

\subsubsection{Experiment Results on Other Datasets} \label{app:other_exp}
We further evaluate the proposed method on the traffic and M4 datasets. Experiment results are shown in Table  Table \ref{tab:m4}. According to experiment results of the M4 dataset, which contains the results for yearly, quarterly, and monthly collected univariate marketing data, we can find that the proposed UDA also outperforms other state-of-the-art deep forecasting models for nonstationary time series forecasting.

\begin{table}[h]
\renewcommand{\arraystretch}{0.9}
\caption{Experiment results on M4 dataset.}
\resizebox{\textwidth}{!}{%
\begin{tabular}{ccccccccccc}
\hline
\multicolumn{2}{c}{models}                                                   & UDA            & Koopa          & SAN    & DLinear & N-Transformer & RevIN  & MICN   & TimeNets & WITRAN         \\ \hline
\multicolumn{1}{c|}{\multirow{3}{*}{Yearly}}      & \multicolumn{1}{c|}{sMAPE} & \textbf{13.357} & 13.761         & 14.631 & 14.312  & 13.817        & 15.04  & 14.759 & 13.544   & 13.648         \\
\multicolumn{1}{c|}{}                           & \multicolumn{1}{c|}{MASE}  & \textbf{2.987}  & 3.049          & 3.254  & 3.096   & 3.054         & 3.091  & 3.362  & 3.030    & 3.053          \\
\multicolumn{1}{c|}{}                           & \multicolumn{1}{c|}{OWA}   & \textbf{0.713}  & 0.732          & 0.779  & 0.752   & 0.734         & 0.964  & 0.796  & 0.724    & 0.729          \\ \hline
\multicolumn{1}{c|}{\multirow{3}{*}{Quarterly}} & \multicolumn{1}{c|}{sMAPE} & \textbf{10.037} & 10.405         & 11.532 & 10.493  & 11.882        & 12.226 & 11.349 & 10.117   & 10.453         \\
\multicolumn{1}{c|}{}                           & \multicolumn{1}{c|}{MASE}  & \textbf{1.114}  & 1.154          & 1.270  & 1.169   & 1.195         & 1.311  & 1.285  & 1.122    & 1.165          \\
\multicolumn{1}{c|}{}                           & \multicolumn{1}{c|}{OWA}   & \textbf{0.825}  & 0.855          & 0.945  & 0.864   & 0.986         & 0.971  & 0.942  & 0.831    & 0.861          \\ \hline
\multicolumn{1}{c|}{\multirow{3}{*}{Monthly}}   & \multicolumn{1}{c|}{sMAPE} & \textbf{12.737} & 12.89          & 13.985 & 13.291  & 14.181        & 14.629 & 13.847 & 12.817   & 13.302         \\
\multicolumn{1}{c|}{}                           & \multicolumn{1}{c|}{MASE}  & \textbf{0.928}  & 0.939          & 1.114  & 0.975   & 1.049         & 1.071  & 1.027  & 0.933    & 0.979          \\
\multicolumn{1}{c|}{}                           & \multicolumn{1}{c|}{OWA}   & \textbf{0.934}  & 0.945          & 1.145  & 0.978   & 1.048         & 1.147  & 1.024  & 0.939    & 0.980          \\ \hline
\multicolumn{1}{c|}{\multirow{3}{*}{Others}}    & \multicolumn{1}{c|}{sMAPE} & \textbf{4.872}  & 4.894          & 5.281  & 5.079   & 6.404         & 6.915  & 6.02   & 5.058    & 6.276          \\
\multicolumn{1}{c|}{}                           & \multicolumn{1}{c|}{MASE}  & 3.115           & 3.076          & 3.427  & 3.567   & 3.442         & 4.122  & 4.127  & 3.247    & \textbf{3.039} \\
\multicolumn{1}{c|}{}                           & \multicolumn{1}{c|}{OWA}   & 0.974           & \textbf{0.951} & 1.049  & 1.062   & 1.134         & 1.448  & 1.239  & 0.998    & 1.059          \\ \hline
\multicolumn{1}{c|}{\multirow{3}{*}{Average}}   & \multicolumn{1}{c|}{sMAPE} & \textbf{11.838} & 12.093         & 13.03  & 12.443  & 13.295        & 13.141 & 13.066 & 11.948   & 12.346         \\
\multicolumn{1}{c|}{}                           & \multicolumn{1}{c|}{MASE}  & \textbf{1.483}  & 1.515          & 1.681  & 1.543   & 1.578         & 1.694  & 1.674  & 1.502    & 1.524          \\
\multicolumn{1}{c|}{}                           & \multicolumn{1}{c|}{OWA}   & \textbf{0.849}  & 0.868          & 1.007  & 0.887   & 0.926         & 1.245  & 1.351  & 0.859    & 0.878          \\ \hline
\end{tabular}}
\label{tab:m4}
\end{table}

\subsection{Sensitive Analysis} \label{app:sensitive}
\begin{figure}
  \centering
\subfigure[]{
    \includegraphics[width=0.23\textwidth]{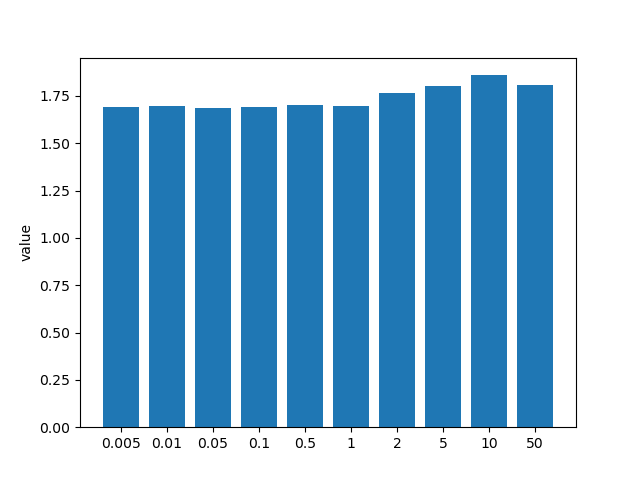}
    \label{fig:sub1}
  }
  \hfill
  \subfigure[]{
    \includegraphics[width=0.23\textwidth]{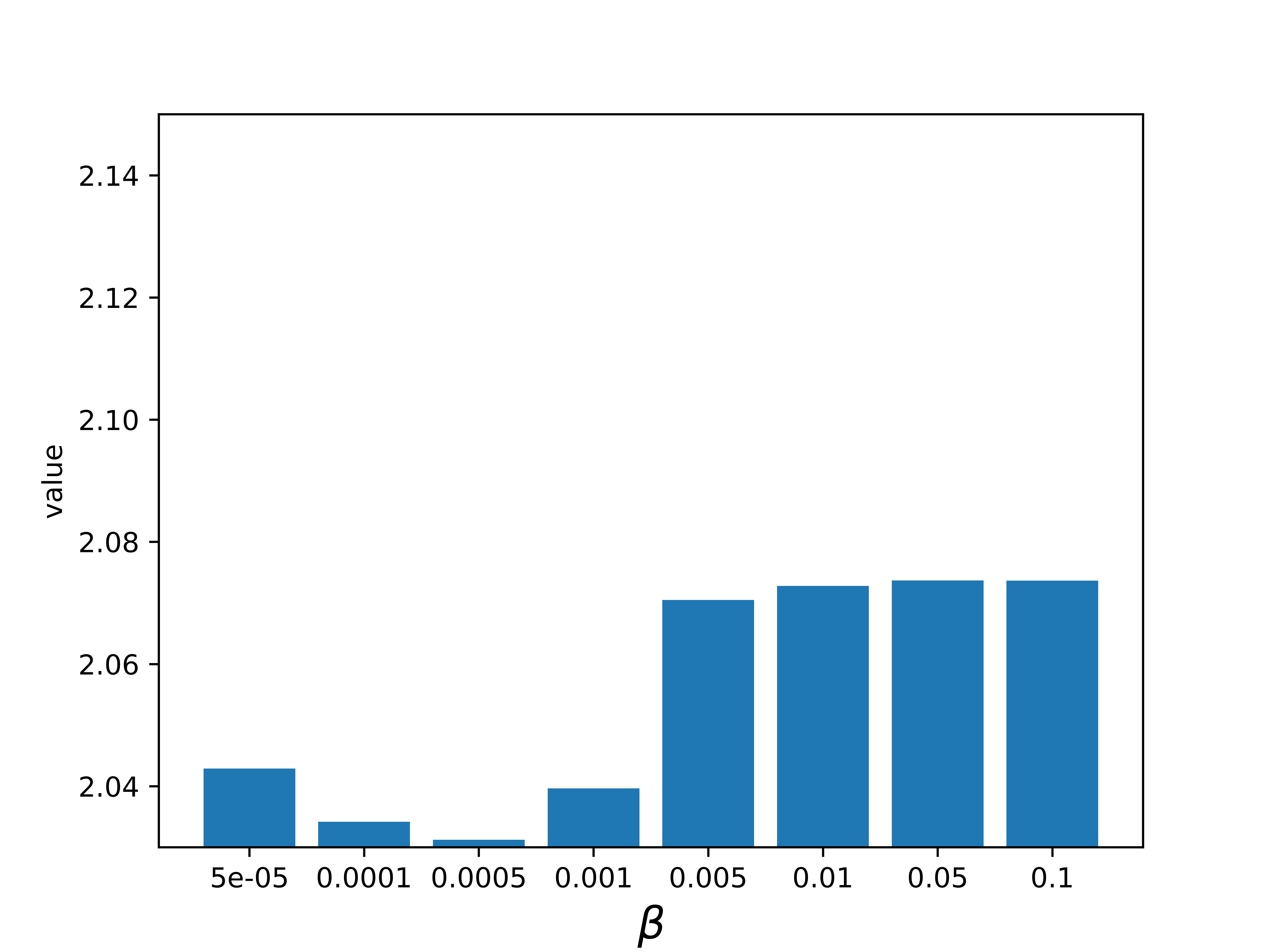}
    \label{fig:sub1}
  }
  \hfill
  \subfigure[]{
    \includegraphics[width=0.23\textwidth]{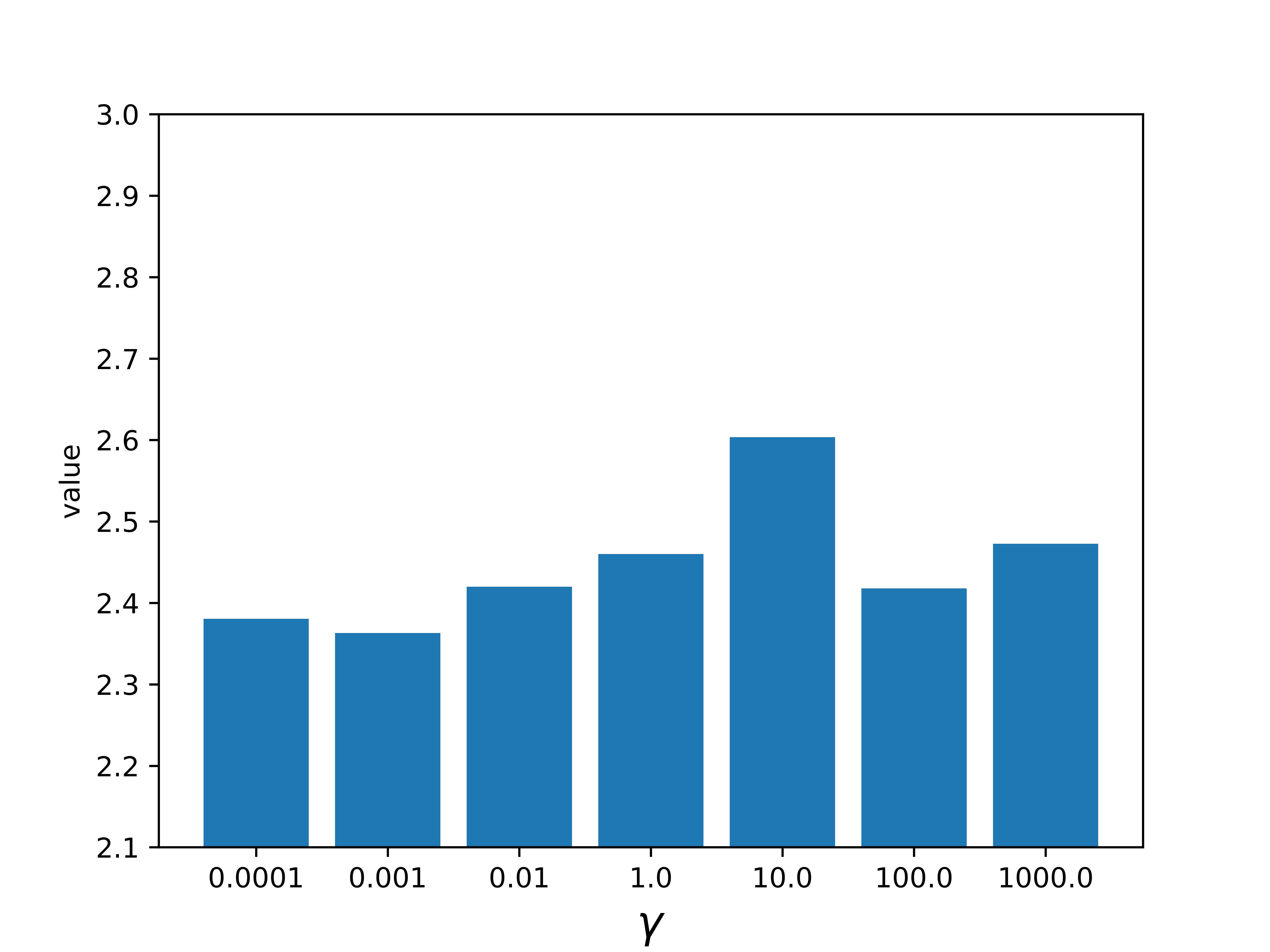}
    \label{fig:sub2}
  }
  \hfill
  \subfigure[]{
    \includegraphics[width=0.23\textwidth]{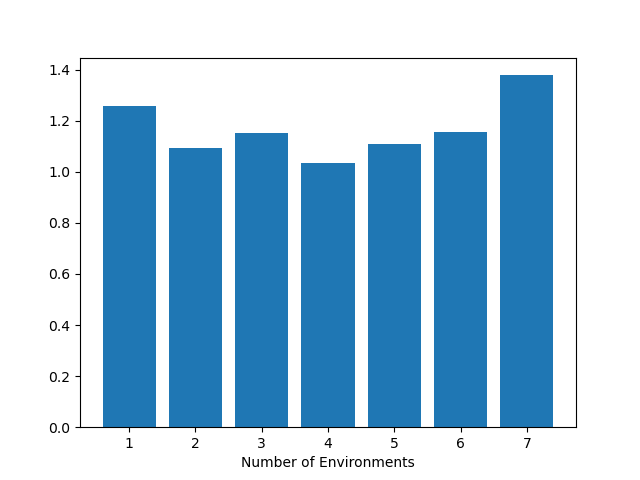}
    \label{fig:sub3}
  }
  \caption{Experiment results of different values of $\alpha, \beta, \gamma$, and prior number of environments }
  \label{fig:sensitive}
\end{figure}

We further try different values of $\alpha,
\beta, \gamma$, and the number of prior environments, which are shown in  Figure \ref{fig:sensitive} (a),(b),(c), and (d), respectively. According to Figure \ref{fig:sensitive} (a)(b)(c), we can find that the experiment results are stables in a specific area of the values of hyperparameters. In our practical implementation, we let the number of latent environments be 4. Since the value of latent environments is considered to be a hyper-parameter, we try different values of the latent environments, which are shown in Figure \ref{fig:sensitive} (d). According to the experiment results, we can find that the experiment results vary with the values of latent environment, reflecting the importance of suitable prior.


\subsection{Ablation Study}
\label{app:ablation}
\vspace{-3mm}
\begin{figure*}[h]
  \centering
\includegraphics[width=\textwidth]{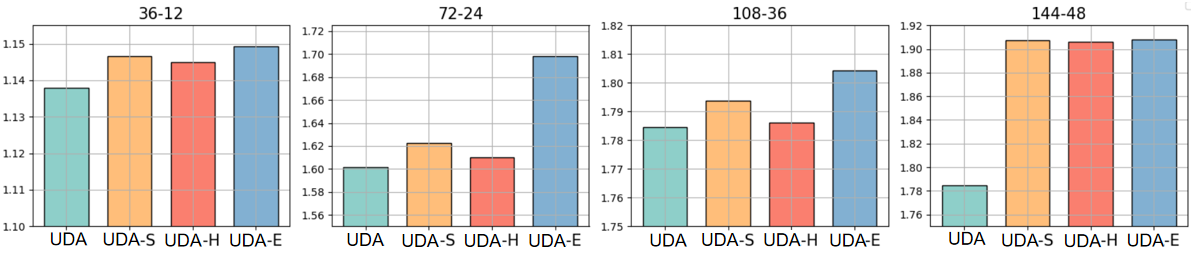}
\vspace{-7mm}
  \caption{Ablation study on the different predict forecast lengths of ILI dataset. we explore the impact of different loss terms.} 
  \label{fig:ablation} 
  \vspace{-4mm}
\end{figure*}


Ablation study of UDA-sh are shown in Table \ref{tab:IDEA_sh}. According to the experiment results, we can draw the following conclusions: 
1) The performance of the standard UDA and the UDA-Sh are similar, this is because the prediction of $x_{1:t}$ and $x_{t+1:T}$ share the same decoding process. 
2) We also find that the performance of UDA is slightly better than that of UDA-Sh in most of the forecasting tasks, reflecting that the model with more parameters may improve the model performance.

\begin{table}[h]
\centering
\caption{Experiment results of UDA and UDA-Sh}
\renewcommand{\arraystretch}{0.70}
\label{tab:IDEA_sh}
\begin{tabular}{@{}c|ccccccccc@{}}
\toprule
                          & \textbf{} & \multicolumn{2}{c}{\textbf{36-12}} & \multicolumn{2}{c}{\textbf{72-24}} & \multicolumn{2}{c}{\textbf{144-48}} & \multicolumn{2}{c}{\textbf{216-72}} \\ \midrule
Dataset                   & \textbf{} & MSE              & MAE             & MSE              & MAE             & MSE              & MAE              & MSE              & MAE              \\ \midrule
\multirow{2}{*}{ECL}      & UDA-Sh   & 0.123            & \textbf{0.215}  & 0.131            & 0.257           & 0.124            & 0.228            & 0.132            & \textbf{0.187}   \\
                          & UDA      & \textbf{0.114}   & 0.216           & \textbf{0.121}   & \textbf{0.22}   & \textbf{0.122}   & \textbf{0.224}   & \textbf{0.131}   & \textbf{0.187}   \\ \midrule
\multirow{2}{*}{ILI}      & UDA-Sh   & 1.241            & 0.711           & 1.69             & 0.814           & 1.805            & \textbf{0.866}   & 1.934            & 0.934            \\
                          & UDA      & \textbf{1.218}   & \textbf{0.694}  & \textbf{1.68}    & \textbf{0.809}  & \textbf{1.792}   & 0.869            & \textbf{1.883}   & \textbf{0.926}   \\ \midrule
\multirow{2}{*}{Weather}  & UDA-Sh   & 0.074            & 0.093           & 0.099            & 0.136           & 0.121            & 0.161            & 0.14             & 0.193            \\
                          & UDA      & \textbf{0.072}   & \textbf{0.09}   & \textbf{0.098}   & \textbf{0.13}   & \textbf{0.115}   & \textbf{0.158}   & \textbf{0.136}   & \textbf{0.187}   \\ \midrule
\multirow{2}{*}{Exchange} & UDA-Sh   & \textbf{0.014}   & 0.075           & 0.024            & 0.103           & 0.043            & 0.143            & 0.065            & \textbf{0.177}   \\
                          & UDA      & \textbf{0.014}   & \textbf{0.074}  & \textbf{0.023}   & \textbf{0.102}  & \textbf{0.042}   & \textbf{0.141}   & \textbf{0.065}   & 0.18             \\ \midrule
\multirow{2}{*}{ETTh1}    & UDA-Sh   & 0.292            & \textbf{0.344}  & \textbf{0.299}   & 0.354           & 0.355            & 0.39             & 0.375            & 0.395            \\
                          & UDA      & \textbf{0.291}   & 0.345           & 0.3              & \textbf{0.353}  & \textbf{0.338}   & \textbf{0.38}    & \textbf{0.367}   & \textbf{0.388}   \\ \midrule
\multirow{2}{*}{ETTh2}    & UDA-Sh   & \textbf{0.14}    & \textbf{0.236}  & \textbf{0.172}   & \textbf{0.26}   & 0.236            & \textbf{0.306}   & 0.288            & 0.344            \\
                          & UDA      & \textbf{0.141}   & \textbf{0.236}  & 0.173            & \textbf{0.26}   & \textbf{0.233}   & \textbf{0.306}   & \textbf{0.262}   & \textbf{0.327}   \\ \bottomrule
\end{tabular}
\end{table}

\section{Model Efficiency}\label{app:eff}
\begin{figure}[h]
  \centering
  \subfigure[Weather]{
    \includegraphics[width=0.45\textwidth]{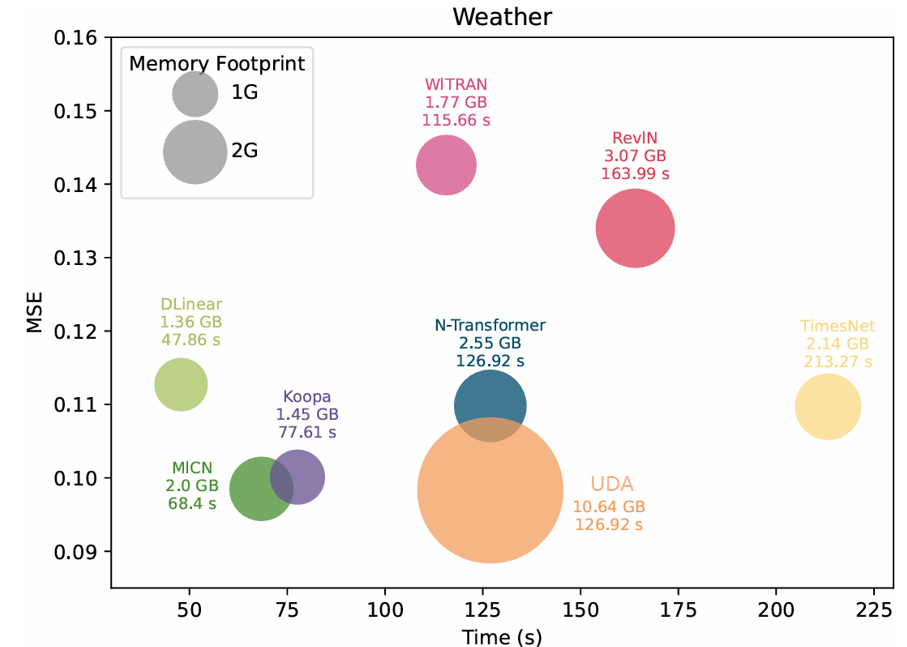}
    \label{fig:subfig1}
  }
  \hfill
  \subfigure[Exchange]{
    \includegraphics[width=0.44\textwidth]{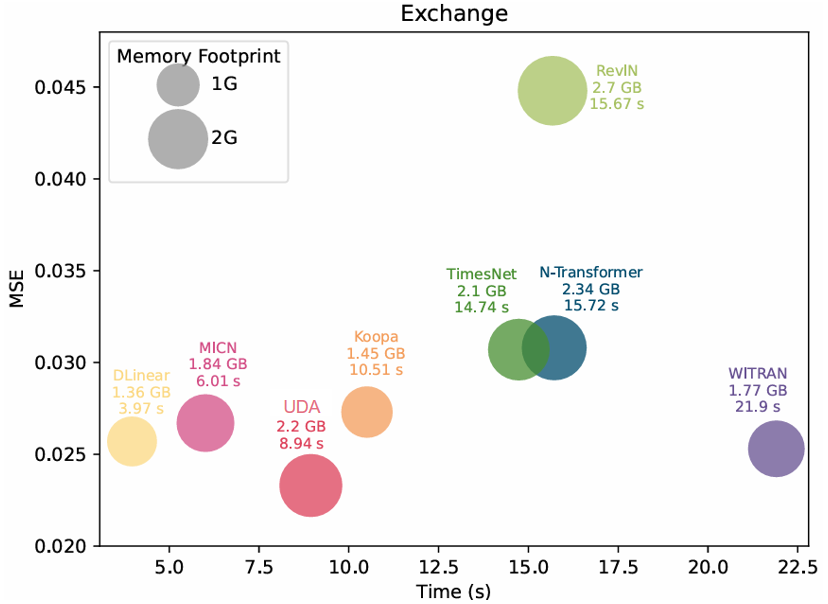}
    \label{fig:subfig2}
  }
  \caption{Model efficiency comparison. Training time and memory footprint are recorded with the }
  \label{fig:effi}
\end{figure}

Following \cite{liu2023koopa} We conduct model efficiency comparasion from three perspectives:  forecasting
performance, training speed, and memory footprint, which is shown in Figure \ref{fig:effi}. Compared with other models for nonstationary time-series forecasting, we can find that the proposed \textbf{UDA} model enjoys the best high model performance and model efficiency, this is because our \textbf{UDA} is built on MLP-based neural architecture. Compared with other methods like MICN and DLinear, our method achieves a weaker model efficiency, this is because our model needs to model the latent-variable-wise prior. 

\section{Implementation Details} \label{app:implementation}
We summarize our network architecture below and describe it in detail in Table \ref{tab:architecture}.
\begin{table}[h]
\renewcommand{\arraystretch}{0.65}
\small
\centering
\caption{Architecture details. BS: batch size, T: length of time series, LeakyReLU: Leaky Rectified Linear Unit, $|\rvx_t|$: the dimension of $\rvx_t$.}
\label{tab:architecture}
\begin{tabular}{@{}c|cc@{}}
\toprule
\textbf{Configuration}       & \multicolumn{1}{c|}{\textbf{Description}}                            & \textbf{Output}         \\ \midrule
1. $\psi_s$                    & \multicolumn{1}{c|}{Stationary Latent Variable Encoder}              &                         \\ \midrule
input:$x_{1:t}$              & \multicolumn{1}{c|}{Observed time series}                            & BS $\times t \times |\rvx_t|$  \\
Permute                      & \multicolumn{1}{c|}{Matrix Transpose}                                & BS $\times$ $|\rvx_t|$ $\times t$   \\
Dense                        & \multicolumn{1}{c|}{384 neurons,LeakyReLU}                           & BS $\times$ $n_s$ $\times 384$   \\
Dense                        & \multicolumn{1}{c|}{t neurons}                                       & BS $\times$ $n_s$ $\times t$   \\
Permute                      & \multicolumn{1}{c|}{Matrix Transpose}                                & BS $\times t $  $\times n_s$      \\ \midrule
2. $T_s$                       & \multicolumn{1}{c|}{Stationary Latent Variable Prediction Module}    &                         \\ \midrule
Input:$z^s_{1:t}$            & \multicolumn{1}{c|}{Stationary Latent Variables}                     &    BS $\times t$ $\times n_s$                        \\
Permute                      & \multicolumn{1}{c|}{Matrix Transpose}                                & BS $\times n_s$ $\times$ t              \\
Dense                        & \multicolumn{1}{c|}{384 neurons,LeakyReLU}                           & BS $\times n_s$ $\times$ 384           \\
Dense                        & \multicolumn{1}{c|}{T-t neurons}                                     & BS $\times n_s$ $\times (T-t)$          \\
Permute                      & \multicolumn{1}{c|}{Matrix Transpose}                                & BS $\times (T-t)$  $\times n_s$         \\ \midrule
3.$\psi_e$                     & \multicolumn{1}{c|}{Nonstationary Latent Variable Encoder}           &                         \\ \midrule
input:$x_{1:t}$              & \multicolumn{1}{c|}{Observed time series}                            & Batch Size $\times$ t $\times$ X dimension  \\
Permute                      & \multicolumn{1}{c|}{Matrix Transpose}                                & BS $\times |\rvx_t|$ $\times t$   \\
Dense                        & \multicolumn{1}{c|}{384 neurons,LeakyReLU}                           & BS $\times |\rvx_t|$ $\times 384$\\
Dense                        & \multicolumn{1}{c|}{128 neurons}                                     & BS $\times |\rvx_t|$ $\times 128$ \\
Dense                        & \multicolumn{1}{c|}{384 neurons,LeakyReLU}                           & BS $\times n_e$ $\times 384$   \\
Dense                        & \multicolumn{1}{c|}{t neurons}                                       & BS $\times n_e$ $\times t$     \\
Permute                      & \multicolumn{1}{c|}{Matrix Transpose}                                & BS $\times$ t $\times n_e$      \\ \midrule
4.$T_e$                      & \multicolumn{1}{c|}{Nonstationary Latent Variable Prediction Module} &                         \\ \midrule
Input:$z^e_{1:t}$              & \multicolumn{1}{c|}{Nonstationary Latent Variables}                  &      BS $\times t$  $\times n_e$                   \\
Permute                      & \multicolumn{1}{c|}{Matrix Transpose}                                & BS $\times n_e$  $\times t$              \\
Dense                        & \multicolumn{1}{c|}{384 neurons,LeakyReLU}                           & BS $\times n_e$ $\times 384$            \\
Dense                        & \multicolumn{1}{c|}{T-t neurons}                                     & BS $\times n_e$  $\times (T-t)$             \\
Permute                      & \multicolumn{1}{c|}{Matrix Transpose}                                & BS $\times (T-t)$ $\times n_e$             \\ \midrule
5.$F_x$                        & \multicolumn{1}{c|}{Historical Decoder}                              &                         \\ \midrule
Input:$z^s_{1:t},z^e_{1:t}$ & \multicolumn{1}{c|}{Stationary and nonstationary Latent Variable} & BS$\times$ t $\times n_s$, BS$\times$ t$\times n_e$ \\
Concat                       & \multicolumn{1}{c|}{concatenation}                                   & BS $\times$ t $\times$ ($n_e+n_s$)   \\
Dense                        & \multicolumn{1}{c|}{x dimension neurons}                                   & BS $\times$ t $\times |\rvx_t|$          \\
Permute                      & \multicolumn{1}{c|}{Matrix Transpose}                                & BS $\times |\rvx_t|$ $\times t$            \\
Dense                        & \multicolumn{1}{c|}{384 neurons,RelU}                                & BS $\times |\rvx_t|$ $\times$ 384          \\
Dense                        & \multicolumn{1}{c|}{t neurons}                                       & BS $\times |\rvx_t|$ $\times t$            \\
Permute                      & \multicolumn{1}{c|}{concatenation}                                   & BS $\times$ t  $\times |\rvx_t|$           \\ \midrule
6.$F_y$                        & \multicolumn{1}{c|}{Future Predictor}                                                       &                         \\ \midrule
Input:$z^s_{t+1:T},z^e_{t+1:T}$ & \multicolumn{1}{c|}{Stationary and Nonstationary Latent Variable} & BS $\times (T-t)$ $\times n_s$ ,BS$\times (T-t)$ $\times n_e$ \\
Concat                       & \multicolumn{1}{c|}{concatenation}                                   & BS $\times (T-t)$ $\times$ ($n_e+n_s$)   \\
Dense                        & \multicolumn{1}{c|}{x dimension neurons}                                   & BS $\times (T-t)$ $\times |\rvx_t|$          \\
Permute                      & \multicolumn{1}{c|}{Matrix Transpose}                                & BS $\times |\rvx_t|$ $\times (T-t)$            \\
Dense                        & \multicolumn{1}{c|}{384 neurons,LeakyReLU}                           & BS $\times |\rvx_t|$ $\times 384$           \\
Dense                        & \multicolumn{1}{c|}{T-t neurons}                                     & BS  $\times |\rvx_t|$ $\times (T-t)$            \\
Permute                      & \multicolumn{1}{c|}{Matrix Transpose}                                & BS $\times (T-t)$$\times |\rvx_t|$          \\ \midrule
{ 7.$r$} & \multicolumn{1}{c|}{Modular Prior Networks}                       &                         \\ \midrule
Input:$z^s_{1:T}$ or $z^e_{1:T}$               & \multicolumn{1}{c|}{Latent Variables}                  & BS $\times$ $(n_*+1)$       \\
Dense                        & \multicolumn{1}{c|}{128 neurons,LeakyReLU}                           & $(n_*+1)\times 128$             \\
Dense                        & \multicolumn{1}{c|}{128 neurons,LeakyReLU}                           & 128$\times$128                 \\
Dense                        & \multicolumn{1}{c|}{128 neurons,LeakyReLU}                           & 128$\times$128                 \\
Dense                        & \multicolumn{1}{c|}{1 neuron}                                        & BS $\times$1                \\
JacobianCompute              & \multicolumn{1}{c|}{Compute log ( det (J))}                          & BS              \\ \bottomrule
\end{tabular}%



\end{table}

\end{document}